\documentclass[journal]{IEEEtai}

\usepackage[colorlinks,urlcolor=blue,linkcolor=blue,citecolor=blue]{hyperref}

\usepackage{color,array}

\usepackage{graphicx}
\usepackage[utf8]{inputenc} 
\usepackage{amsmath,amsfonts}
\usepackage{algorithmic}
\usepackage{graphicx}
\usepackage{multicol,multirow,rotating,booktabs,float}
\usepackage{tabularx}
\usepackage{supertabular}
\usepackage[export]{adjustbox}
\usepackage{textcomp}
\usepackage{fancyhdr}
\usepackage{xcolor}
\usepackage{setspace}
\usepackage[ruled, linesnumbered, noline]{algorithm2e}

\usepackage{booktabs}
\usepackage{pifont}   
\usepackage[table]{xcolor}

\makeatletter
\disable@package@load{burl}{}
\makeatother

\begin{document}

\title{Diffusion-Assisted Distillation for Self-Supervised Graph Representation Learning with MLPs}

\author{
    Seong Jin Ahn,\, 
    Myoung-Ho Kim\,
\IEEEcompsocitemizethanks{%
    \IEEEcompsocthanksitem S.~J.~Ahn and M.-H.~Kim are with the School of Computing, KAIST, Daejeon 34141, South Korea (e-mail: sja1015@kaist.ac.kr, mhkim@kaist.ac.kr).
  }%
\thanks{This work was supported in part by the MSIT(Ministry of Science and ICT),
Korea, under the ITRC(Information Technology Research Center) support program(IITP-2024-2020-0-01795) supervised by the IITP(Institute of Information \& Communications Technology Planning \& Evaluation) and in part by the Samsung Electronics (IO201209-07901-01).}

\thanks{This work is the accepted author manuscript of the article:
S. J. Ahn et al., "Diffusion-Assisted Distillation for Self-Supervised Graph Representation Learning with MLPs," 
\textit{IEEE Transactions on Artificial Intelligence}, 2025. 
DOI: \href{https://doi.org/10.1109/TAI.2025.3598791}{10.1109/TAI.2025.3598791}}
}

\markboth{Diffusion-Assisted Distillation for Self-Supervised Graph Representation Learning with MLPs}
{Ahn \MakeLowercase{\textit{et al.}}: Diffusion-Assisted Distillation for Self-Supervised Graph Representation Learning with MLPs}

\maketitle

\begin{abstract}
For large-scale applications, there is growing interest in replacing Graph Neural Networks (GNNs) with lightweight Multi-Layer Perceptrons (MLPs) via knowledge distillation.
However, distilling GNNs for self-supervised graph representation learning into MLPs is more challenging.
This is because the performance of self-supervised learning is more related to the model's inductive bias than supervised learning. 
This motivates us to design a new distillation method to bridge a huge capacity gap between GNNs and MLPs in self-supervised graph representation learning.
In this paper, we propose \textbf{D}iffusion-\textbf{A}ssisted \textbf{D}istillation for \textbf{S}elf-supervised \textbf{G}raph representation learning with \textbf{M}LPs (DAD-SGM).
The proposed method employs a denoising diffusion model as a teacher assistant to better distill the knowledge from the teacher GNN into the student MLP. 
This approach enhances the generalizability and robustness of MLPs in self-supervised graph representation learning.
Extensive experiments demonstrate that DAD-SGM effectively distills the knowledge of self-supervised GNNs compared to state-of-the-art GNN-to-MLP distillation methods.
Our implementation is available at \url{https://github.com/SeongJinAhn/DAD-SGM}.
\end{abstract}

\begin{IEEEImpStatement}
This paper presents Diffusion-Assisted Distillation for Self-supervised Graph representation learning with MLPs (DAD-SGM), a novel framework that addresses the performance gap between GNNs and MLPs in self-supervised graph learning. 
Our approach first trains an assistant denoising diffusion model that learns to predict noise from noisy outputs of the GNN teacher.
Using this assistant model, DAD-SGM aligns the density gradients of the teacher and student outputs, facilitating effective knowledge transfer. 
Extensive experiments demonstrate that DAD-SGM consistently surpasses existing distillation methods across diverse datasets. 
Specifically, DAD-SGM provides up to 15\% and 19\% improvement in node classification and link prediction performance over existing GNN-to-MLP distillation methods.
This approach opens up new possibilities for performing self-supervised representation learning on large-scale graphs.
\end{IEEEImpStatement}


\section{Introduction}
\IEEEPARstart{G}raph Neural Networks (GNNs) have emerged as a powerful tool for analyzing non-Euclidean graph data.
GNNs capture complex graph structures through message-passing, wherein nodes exchange messages with their neighbors. 
This message-passing mechanism provides a spatial inductive bias that is beneficial for analyzing graph-structured data.
However, as graph sizes increase, this message-passing approach incurs significant computational overhead~\cite{tian2022nosmog}. 
To address this scalability challenge, researchers have explored replacing GNNs with more lightweight Multilayer Perceptrons (MLPs).
While MLPs lack the structural bias provided by message-passing, they offer significantly greater scalability. 
To combine the strengths of both GNNs and MLPs, researchers have increasingly turned to knowledge distillation~\cite{hinton2015distilling}, a promising technique for transferring the task-specific knowledge from GNNs into MLPs.
For semi-supervised node classification, GLNN \cite{zhang2022graphless} minimizes the Kullback-Leibler divergence between the logits of a GNN teacher and an MLP student. 
FF-G2M \cite{wu2023extracting} decomposes the knowledge of the teacher model into high- and low-frequency signals and then encourages the student model to mimic both of them.
For link prediction, LLP \cite{guo2023linkless} distills relation-specific knowledge centered around each (anchor) node to the student MLP.
These GNN-to-MLP distillation methods enable MLPs to achieve comparable performance to GNNs in their target tasks.

\begin{figure}[t]
    \begin{minipage}[b]{0.45\textwidth}
      \centerline{\includegraphics[width=8cm]{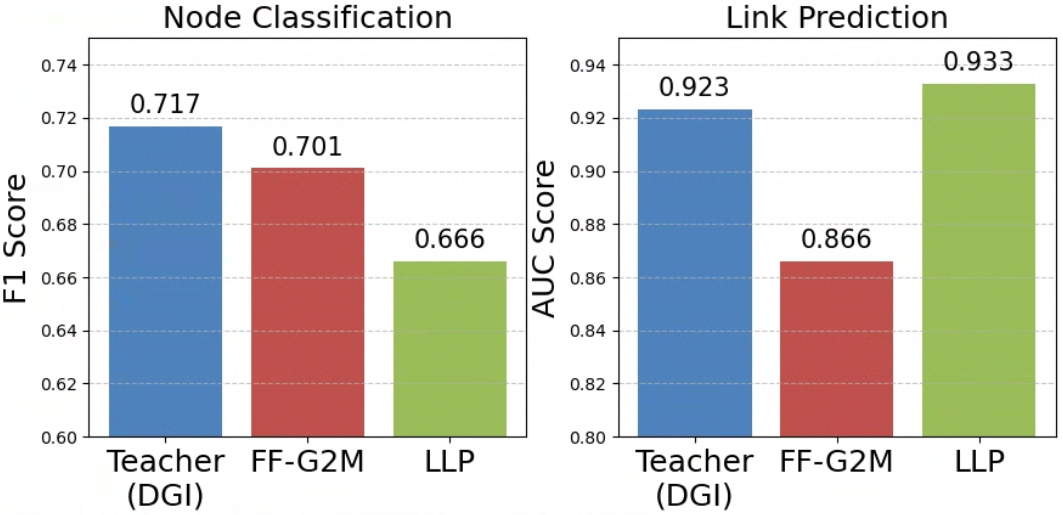}}
      \centerline{(a)} \medskip
    \end{minipage}
    \begin{minipage}[b]{0.45\textwidth}
      \centerline{\includegraphics[width=8.5cm]{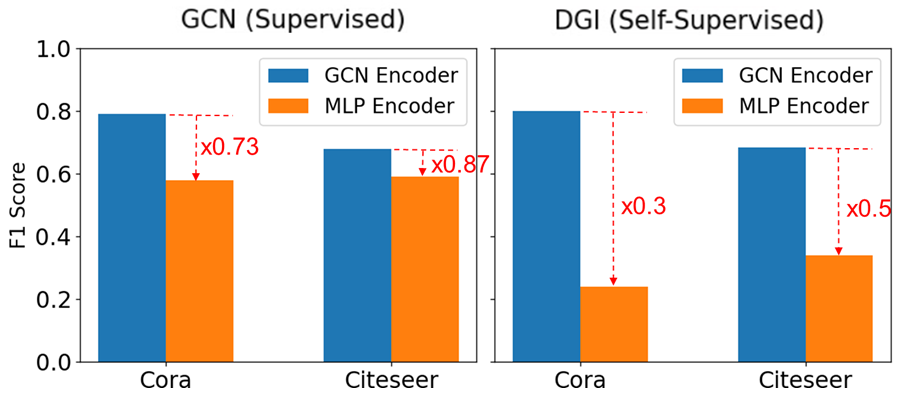}}
      \centerline{(b)}\medskip
    \end{minipage}
    \caption{(a) The node classification and link prediction performance of two GNN-to-MLP distillation methods (FF-G2M and LLP) on the Citeseer dataset, and (b) The node classification performance of a 1-layer GCN and 3-layer MLP in supervised and unsupervised settings.}
    \label{fig1}
\end{figure}

Self-supervised graph representation learning (SSGRL) has emerged as a crucial paradigm for analyzing graph-structured data without label information.  
Its main objective is to encode the nodes into low-dimensional representations that can be used effectively in various downstream tasks. 
With the development of GNNs, a variety of GNN-based approaches have been proposed for SSGRL.
For example, DGI \cite{veličković2018deep} contrasts node representations with a global summary to capture consistent yet discriminative features, while CCA-SSG \cite{zhang2021canonical} maximizes the correlation among node representations generated by two different graph augmentations.
Despite their success, these self-supervised GNNs rely heavily on message passing, which poses scalability challenges when applied to large graphs.  
We expect the effective knowledge transfer from self-supervised GNNs to MLPs to offer a promising way to achieve both scalability and efficiency in SSGRL.

However, current GNN-to-MLP distillation methods are task-specific and often fail to preserve the task-agnostic knowledge learned by self‑supervised teachers.
As shown in Figure \ref{fig1}(a), FF‑G2M retains most of a DGI teacher’s node classification accuracy but degrades its link prediction performance.
LLP exhibits the opposite behavior, maintaining link prediction performance while losing node classification accuracy.
This inconsistency stems from the fact that model capacity plays a more critical role in capturing task-agnostic knowledge than task-specific knowledge ~\cite{ashihara2022deep,tian2019contrastive,zhong2023contrastive}.    
As illustrated in Figure~\ref{fig1}(b), the capacity gap between a heavy GNN teacher and a lightweight MLP student is also pronounced in self-supervised settings.
Therefore, it becomes more challenging to transfer the task-agnostic knowledge in self-supervised GNNs to lightweight MLPs.
Bridging this enlarged teacher–student gap requires a more sophisticated strategy.
In other domains, several researchers have leveraged teacher assistant (TA) models to bridge the capacity gap between teacher and student models~\cite{mirzadeh2020improved,yoon2023takdsr,li2024tas,han2024amd}.  
The TA models are intermediate models that facilitate smoother knowledge transfer between large teachers and lightweight students.
A well-designed TA should (i) preserve the essential knowledge of the teacher and (ii) transfer it in a form that is accessible to the student.

In this paper, we propose a new GNN-to-MLP distillation framework, named \textbf{D}iffusion-\textbf{A}ssisted \textbf{D}istillation for \textbf{S}elf-supervised \textbf{G}raph representation learning with \textbf{M}LPs (DAD-SGM).
Our key idea is to employ an MLP-based denoising diffusion model as a teacher assistant (TA) between the teacher GNN and the student MLP.
Denoising Diffusion Models (DDMs) have recently emerged as powerful generative models, achieving state-of-the-art results in multiple domains.
Their success is largely attributed to their ability to predict the noise in data, enabling them to capture fine-grained knowledge and improve robustness to noisy inputs
 \cite{ho2020denoising,song2020denoising}. 
Our proposed TA leverages this denoising diffusion process to effectively retain the teacher’s key knowledge, while its MLP-based architecture ensures compatibility for efficient knowledge transfer to the student.

Specifically, DAD-SGM consists of two stages.
The first stage trains the MLP-based DDM assistant.
The assistant model learns a noise prediction function $\epsilon_\phi$ that estimates noise from noisy node representations of the teacher GNN.
The predicted noise captures the fine-grained information of the teacher's node representations.
The second stage involves distilling the teacher GNN into a student MLP using the trained assistant.
Aligning the estimated noise between teacher and student outputs makes the student MLP more closely approximate the teacher's representation.
To achieve this, we calculate the estimated noise for both the teacher and student representations using the trained noise prediction function $\epsilon_\phi$ from the assistant model.
Then, we update the student’s parameters to minimize the discrepancy between these scores.
During inference, DAD-SGM relies solely on the student MLP, ensuring computational efficiency in practical deployments.

Extensive experiments on eight benchmark datasets, including five homophilic graphs (i.e., graphs where connected nodes are similar) and three heterophilic graphs (i.e., graphs where connected nodes are opposite), demonstrate that DAD-SGM consistently outperforms state-of-the-art GNN-to-MLP knowledge distillation methods.
DAD-SGM achieves notable accuracy improvements compared to existing methods, up to 15\% and 19\% in node classification and link prediction performance, without compromising inference speed.
Additionally, we show that node representations produced by our work achieve robustness comparable to the teacher GNN.
Our further analysis demonstrates that our TA successfully bridges the teacher-student gap in self-supervised graph representation learning, compared to other TA candidates or graph data augmentation strategies.
We expect DAD-SGM to effectively enable fast and accurate self-supervised representation learning in large-scale graph scenarios.

Our contributions are as follows:
\begin{enumerate}
    \item 
    We pioneer a new research direction for distilling task-agnostic knowledge from self-supervised GNNs to MLPs.  
    
    \item 
    We propose DAD-SGM, a novel framework that leverages an MLP-based denoising diffusion model as a teacher assistant to facilitate knowledge transfer from GNNs to MLPs.

    \item 
    We empirically demonstrate that DAD-SGM enhances the generalization ability of MLPs in self-supervised learning. 
    Ablation studies further verify that our diffusion-based TA narrows the teacher–student gap more effectively than alternative TA designs or graph-augmentation strategies.
\end{enumerate}

\section{Related Work}
\subsection{Self-Supervised Graph Representation Learning.}
Self-supervised graph representation learning transforms raw graph data into low-dimensional representations without requiring label information.
GNNs have achieved notable success in self-supervised graph representation learning.
GAE \cite{kipf2016variational} attempts to predict missing edges using a GNN encoder and an inner product decoder.
GRACE \cite{zhu2020deep} maximizes the agreement between two node-level graph views.
DGI \cite{veličković2018deep} contrasts node representations and the high-level summary of graphs. 
In addition, there are contrastive learning methods that learn node representations without relying on negative samples.
GBT \cite{bielak2022graph} reduces the cross-correlation among features to learn node representations. 
CCA-SSG \cite{zhang2021canonical} maximizes the correlation between two augmented views while decorrelating feature dimensions of a single view's representation.
There are attempts to analyze graphs with low homophily, where connected nodes have different characteristics.
SELENE \cite{zhong2022unsupervised} and HGRL \cite{chen2022towards} learn node representations of graphs with low homophily using the Barlow-Twins \cite{zbontar2021barlow} and BYOL \cite{grill2020bootstrap}, respectively.

\subsection{GNN-to-MLP Knowledge Distillation.} 
While graph neural networks (GNNs) have shown remarkable performance in various graph analysis tasks, their dependence on message passing results in delays due to fetching neighbors, hindering their use in large-scale applications \cite{zhang2020agl,jia2020redundancy}.
To resolve this, several GNN-to-MLP knowledge distillation methods have been proposed.
They aim to transfer GNN knowledge into computationally efficient MLPs that do not require explicit message passing  \cite{zhang2022graphless,tian2022nosmog}.
GLNN \cite{zhang2022graphless} first distills the knowledge of trained GNNs to MLPs.
NOSMOG \cite{tian2022nosmog} improves GLNN by injecting positional encoding and structural node similarities.
FF-G2M \cite{wu2023extracting} extracts both low- and high-frequency knowledge from teacher GNNs and injects it into student MLPs.
HGMD \cite{wu2024teach} distills hardness-aware knowledge from teacher GNNs into the corresponding nodes of student MLPs. 
LLP \cite{guo2023linkless} transfers relation-specific knowledge centered around anchors into students for link prediction.
Despite their success, such methods focus on distilling task-specific knowledge (e.g., node classification or link prediction).
Effective GNN-to-MLP distillation for self-supervised graph representation learning remains an under-explored area.

\subsection{Denoising Diffusion Models.}
Denoising diffusion models have recently demonstrated outstanding generative capabilities in computer vision. 
Inspired by non-equilibrium thermodynamics, they learn to reverse a diffusion process and recover original data.
DDPM \cite{ho2020denoising} corrupts images using Gaussian noises in a diffusion process and then recovers them by minimizing a denoising score matching objective.
DDIM \cite{song2020denoising} generalizes the diffusion process of DDPM to a non-Markovian process.
Latent Diffusion Model (LDM) \cite{rombach2022high} reduces the computational cost by performing a diffusion process in latent space.
Several studies have applied DDM to other fields.
EDM \cite{hoogeboom2022equivariant} learns a denoising diffusion model with an equivariant network for 3D molecule generation.
Diffusion-LM \cite{li2022diffusion} trains continuous latent variables with a denoising diffusion process for controllable text generation.
DPM-SNC \cite{jang2023diffusion} learns a joint distribution over known and unknown labels for semi-supervised node classification.
Score distillation sampling (SDS) \cite{pooledreamfusion} has shown great promise in text-to-3D generation by distilling pre-trained text-to-image diffusion models.

\section{Methodology}
Let $\mathcal{G}=(\mathcal{V},\mathcal{E},X)$ be a graph where $\mathcal{V}$ is a set of nodes, $\mathcal{E}$ is a set of edges, and $X=[x_{v_1}, ..., x_{v_{|V|}}]$ is a matrix of node attributes.
Given a pre-trained teacher GNN that encodes nodes in $\mathcal{V}$ to $d$-dimensional node representations $[\textbf{\textit{h}}^\text{(Tea)}_{v_1},...,\textbf{\textit{h}}^\text{(Tea)}_{v_{|\mathcal{V}|}}]$ through self-supervised learning, our work aims to train an MLP model that encodes nodes into $d$-dimensional node representations $[\textbf{\textit{h}}^\text{(Stu)}_{v_1},...,\textbf{\textit{h}}^\text{(Stu)}_{v_{|\mathcal{V}|}}]$ that well preserve the characteristics of the teacher's outputs.
To achieve this, DAD-SGM transfers knowledge from the self-supervised GNN to an MLP by aligning the density gradients of their outputs through an assistant denoising diffusion model.
This process involves a two-stage training process: (i) Training an assistant denoising diffusion model and (ii) Training a student MLP model.

\subsection{First Stage: Training an Assistant Denoising Diffusion Model}
\begin{figure}[t]
   \centerline{\includegraphics[width=9cm]{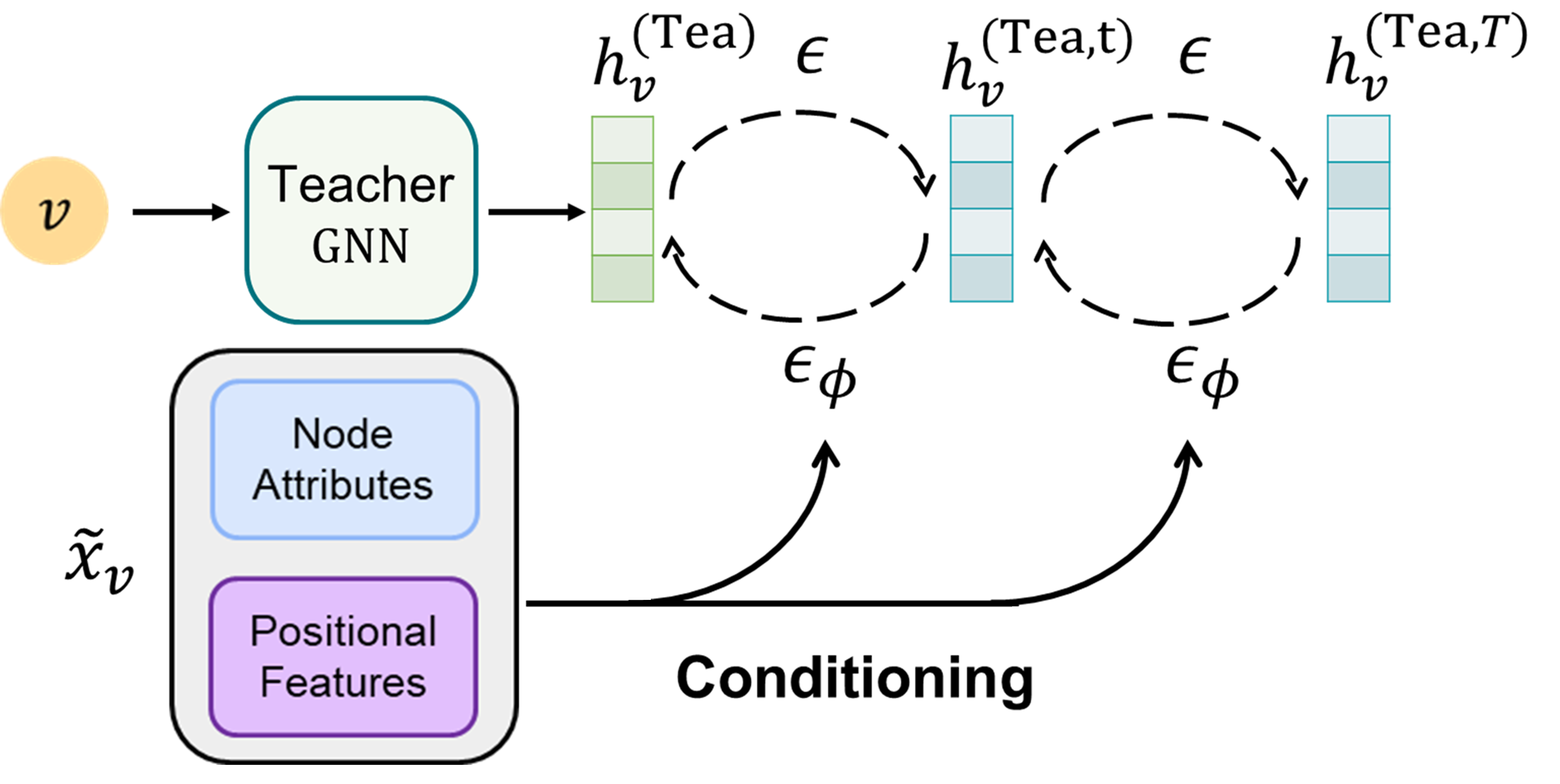}}
    \caption{
        The procedure of training our assistant denoising diffusion model in the first stage.
    }
    \label{fig2}
    \vspace{-1em}
\end{figure}

DAD-SGM begins by training an MLP-based denoising diffusion model parameterized by $\phi$ as a teacher assistant.
The assistant model learns to predict noise from noisy representations of the teacher model with the denoising diffusion process as Figure \ref{fig2}.
This noise prediction $\epsilon_\phi$ is proportional to the gradient of the log probability density \cite{song2020score}.
By accurately predicting the noise, the assistant model can better approximate the output distribution of the teacher model, leading to more effective knowledge transfer.

Let $\textbf{\textit{h}}^\text{(Tea)}_v$ denote the teacher model's representation of node $v$, and $T$ indicates the total diffusion steps.
DAD-SGM follows a non-Markovian generative process as defined in DDIM \cite{song2020denoising}.
For a diffusion time $0\leq t\leq T$, DAD-SGM generates a noisy representation of $v$ at $t$ as:
\begin{equation} 
    \textbf{\textit{h}}_v^{\text{(Tea,t)}} = \sqrt{{\alpha}_t}\text{ }\textbf{\textit{h}}^\text{(Tea)}_v + \sqrt{1-{\alpha}_t}\epsilon,
\label{eq: diff}
\end{equation}
where $\epsilon$ $\sim$$\mathcal{N}(0,\mathbf{I})$ is a Gaussian noise.

{\color{black}{To control the amount of noise added at each step, we adopt a cosine-based scheduling function \cite{nichol2021improved} as Eq.(\ref{eq: schedule}).
This ensures smooth transitions between original and noisy representations by gradually decreasing the weight on the original representations.}}
\begin{equation}
\begin{split}
    f(t) &= \text{cos}^2(\frac{t/T+s}{1+s} \cdot  {\frac{\pi}{2}}),    \\
    \alpha_t &= \prod_{k=1}^t{\frac{f(k)}{f(0)}},
\end{split}
\label{eq: schedule}
\end{equation}
Here, $s$ is an offset that prevents singularities for large $t$.
We set $s$ as 0.001 in this paper.
This scheduling makes $\textbf{\textit{h}}_v^{\text{(Tea,t)}}$ to follow $\mathcal{N}(0,\textbf{I})$.

\begin{figure*}[t]
   \centerline{\includegraphics[width=14cm]{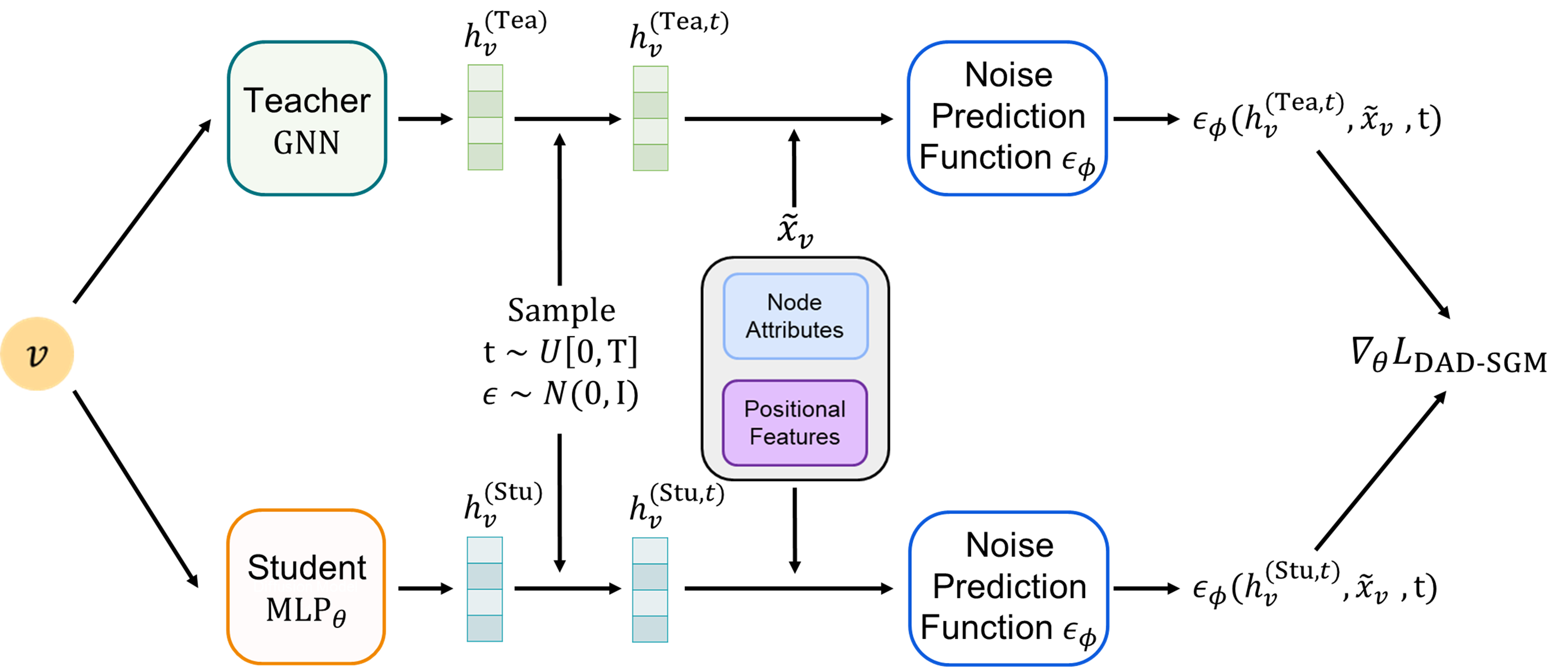}}
    \caption{
        An overview of the proposed DAD-SGM to train its student MLP diffusion model.
    }
    \label{fig3}
\end{figure*}
Then, the assistant model learns to denoise the noisy representations of the teacher model with node attributes and positional features.
This enables the assistant model to capture attribute and structure information in graph-structured data.
For a node $v$, DAD-SGM utilizes the concatenation of node attributes and positional features as
\begin{equation}
    \tilde{\emph{\textbf{x}}}_v = \text{CONCAT}(\emph{\textbf{x}}_v, \emph{\textbf{pf}}_v)
\end{equation}
where $\emph{\textbf{x}}_v$ and $\emph{\textbf{pf}}_v$ are node attributes and positional features of $v$, respectively.
In our implementation, we learn positional features $\emph{\textbf{pf}}_v$ for a node $v$ with the DeepWalk algorithm \cite{perozzi2014deepwalk}.

We encourage the assistant diffusion model to predict the added noise $\epsilon$ using $\tilde{\emph{\textbf{x}}}_v$ for each noisy node representation. 
This is achieved by updating $\phi$ to minimize the difference between the true noise and the estimated noise for each noisy node representation as follows:
\begin{equation}
\begin{split}
    &\nabla_\phi\mathcal{L}_{\text{Assist}}=\\
    &\frac{1}{|\mathcal{V}|} \sum_{v\in\mathcal{V}}\mathbb{E}_{t,\epsilon}\Big[w(t)\Big(\epsilon_\phi(\textbf{\textit{h}}_v^{\text{(Tea,t)}}, \tilde{\textbf{\textit{x}}}_v, t)-\epsilon\Big)\frac{\partial\epsilon_\phi(\textbf{\textit{h}}_v^{\text{(Tea,t)}}, \tilde{\textbf{\textit{x}}}_v, t)}{\partial\phi}\Big]
\end{split}
\end{equation}
where $w(t)=-1/\sqrt{1-\alpha_t}$, $\epsilon\sim\mathcal{N}(0,I)$ is a Gaussian noise, and $\epsilon_\phi$ is an MLP parameterized by $\phi$.
{\color{black}{The gradient $\nabla_\phi\mathcal{L}_{\text{Assist}}$ expresses how the assistant model updates its parameters by weighting the error between the score network’s predicted noise $\epsilon_\phi(\textbf{\textit{h}}_v^{\text{(Tea,t)}}, \tilde{\textbf{\textit{x}}}_v, t)$ and the true noise $\epsilon$.}}
Algorithm \ref{algo1} shows the procedure of training our assistant model.

\RestyleAlgo{ruled}
\begin{algorithm}[t]
  \caption{First Stage: Train an assistant DDM}
  \label{algo1}
  \SetAlgoLined
    $\nabla_\phi\mathcal{L}_\text{Assist}\gets 0$    \\
    
    \For{$v \in \mathcal{V}$}
    {    
        {$\epsilon \sim \mathcal{N}(0,\textbf{I})$}     \\
        $t \sim \mathcal{U}({1,...,T})$  \\
        $\tilde{\textit{\textbf{x}}}_v \gets \text{CONCAT}(\textit{\textbf{x}}_v, \textit{\textbf{pf}}_v)$   \\
        $\textbf{\textit{h}}_v^{\text{(Tea,t)}} = \sqrt{{\alpha}_t}\text{ }\textbf{\textit{h}}^\text{(Tea)}_v + \sqrt{1-{\alpha}_t}\epsilon$ \\

        $\nabla_\phi\mathcal{L}_{\text{Assist}}\leftarrow \nabla_\phi\mathcal{L}_{\text{Assist}}+$ \\
        $\frac{1}{|\mathcal{V}|}\Big[w(t)\Big(\epsilon_\phi(\textbf{\textit{h}}_v^{\text{(Tea,t)}}, \tilde{\textbf{\textit{x}}}_v, t)-\epsilon\Big)\frac{\partial\epsilon_\phi(\textbf{\textit{h}}_v^{\text{(Tea,t)}}, \tilde{\textbf{\textit{x}}}_v, t)}{\partial\phi}\Big]$  \\
    }
  $\textbf{update } \phi$ \\
\end{algorithm}
\RestyleAlgo{ruled}

\RestyleAlgo{ruled}
\begin{algorithm}[t]
  \caption{Second Stage: Train a student MLP}
  \SetAlgoLined
    $\nabla_\theta\mathcal{L}_\text{DAD-SGM}\gets 0$    \\
    
    \For{$v \in \mathcal{V}$}
    {    
        {$\epsilon \sim \mathcal{N}(0,\textbf{I})$}     \\
        $t \sim \mathcal{U}({1,...,T})$  \\
        $\tilde{\textit{\textbf{x}}}_v \gets \text{CONCAT}(\textit{\textbf{x}}_v, \textit{\textbf{pf}}_v)$   \\
        $\textbf{\textit{h}}^\text{(Stu)}_v \gets \text{MLP}_\theta(\tilde{\textit{\textbf{x}}}_v)$
        \\
        $\textbf{\textit{h}}_v^{\text{(Tea,t)}} = \sqrt{{\alpha}_t}\text{ }\textbf{\textit{h}}^\text{(Tea)}_v + \sqrt{1-{\alpha}_t}\epsilon$ \\
        $\textbf{\textit{h}}_v^{\text{(Stu,t)}} = \sqrt{{\alpha}_t}\text{ }\textbf{\textit{h}}^\text{(Stu)}_v + \sqrt{1-{\alpha}_t}\epsilon$ \\
        $\nabla_\theta\mathcal{L}_\text{DAD-SGM}\leftarrow\nabla_\theta\mathcal{L}_\text{DAD-SGM}+$ \\
        $\frac{1}{|\mathcal{V}|}\Big[w(t)\Big(\epsilon_\phi(\textbf{\textit{h}}_v^{\text{(Stu,t)}}, \tilde{\textbf{\textit{x}}}_v, t)-\epsilon_\phi(\textbf{\textit{h}}_v^{\text{(Tea,t)}}, \tilde{\textbf{\textit{x}}}_v, t)\Big){\frac{\partial \textbf{\textit{h}}_v^{\text{(Stu)}}}{\partial\theta}}\Big]$
    }
    \textbf{update} $\theta$
\end{algorithm}
\RestyleAlgo{ruled}

\subsection{Second Stage: Training a student MLP with the Assistant Diffusion Model}
In the second stage, DAD-SGM distills the knowledge of the teacher GNN into the vanilla MLP student by leveraging the estimated noise $\epsilon_\phi$ from the assistant model.
Previous studies have shown that the noise prediction function $\epsilon_\phi$ is proportional to the score function, which represents the gradient of the log probability density \cite{katzirnoise}. 
By matching the estimated noise that serves as the gradient of the teacher and student latent distributions, we guide the student to learn a similar representation space as the teacher.


For each node $v$, the student MLP encodes a node representation $\textbf{\textit{h}}^\text{(Stu)}_v$ based on its node attributes and positional features as
\begin{equation} 
    \textbf{\textit{h}}^\text{(Stu)}_v = \text{MLP}_\theta(\tilde{x}_v),
\label{eq: student}
\end{equation}
where $v\in\mathcal{V}$ is a node, and $\theta$ are trainable parameters of the student MLP.

{\color{black}{The student MLP is trained to align its score estimates with those of the teacher GNN.
To achieve this}}, DAD-SGM generates a noisy representation of $\textbf{\textit{h}}_v^{\text{(Stu)}}$ at a diffusion time $t$ as
\begin{equation} 
    \textbf{\textit{h}}_v^{\text{(Stu,t)}} = \sqrt{{\alpha}_t}\text{ }\textbf{\textit{h}}^\text{(Stu)}_v + \sqrt{1-{\alpha}_t}\epsilon,
\label{eq: student_diff}
\end{equation}
where $\epsilon$ $\sim$$\mathcal{N}(0,\mathbf{I})$ is a Gaussian noise.
Similarly, it produces $\textbf{\textit{h}}_v^{\text{(Tea,t)}}$ as Eq.(\ref{eq: diff}).

To update the student's parameters $\theta$, DAD-SGM minimizes the difference of the predicted noise $\epsilon_\phi$ between noisy teacher representations and noisy student representations as Eq.(\ref{eq7}).
{\color{black}{This equation defines the gradient used to update the student’s parameters by averaging the difference between the student's and teacher's score function outputs.}}
\begin{equation}\label{eq7}
\begin{split}
    &\nabla_\theta\mathcal{L}_\text{DAD-SGM}=\\
    &\frac{1}{|\mathcal{V}|} \sum_{v\in\mathcal{V}}\mathbb{E}_{t,\epsilon}\Big[w(t)\Big(\epsilon_\phi(\textbf{\textit{h}}_v^{\text{(Stu,t)}}, \tilde{\textbf{\textit{x}}}_v, t)-\epsilon_\phi(\textbf{\textit{h}}_v^{\text{(Tea,t)}}, \tilde{\textbf{\textit{x}}}_v, t)\Big){\frac{\partial \textbf{\textit{h}}_v^{\text{(Stu)}}}{\partial\theta}}\Big],\\
\end{split}
\end{equation}
where $w(t)=-1/\sqrt{1-\alpha_t}$.
This formulation utilizes the connection between the score and noise prediction function, expressed as $s_\phi(h,\tilde{x},t)\approx-\epsilon_\phi(h,\tilde{x},t)/\sqrt{1-\alpha_t}$ \cite{zhu2023denoising}.

For inference, DAD-SGM produces student node representations $\textbf{\textit{h}}_v^\text{(Stu)}$ by encoding the concatenation of node attributes and positional features of node $v$ using the student MLP as Eq.(\ref{eq: student}).
This requires the time complexity $O(|\mathcal{V}|d)$ where $d$ is the dimension of node representations.

\begin{table}[t]
  \caption{Statistics of datasets.}
  \label{tab1}
  \centering
  \resizebox{0.5\textwidth}{!}
  {\small
  \begin{tabular}{ c | c c c c c c}
    \toprule
    \multirow{2}{*}{\textbf{Dataset}} & \textbf{\# of} & \textbf{\# of} & \textbf{\# of} & \textbf{\# of} & \textbf{Homophily} \\
     & \textbf{nodes} & \textbf{edges} & \textbf{features} & \textbf{classes} & \textbf{rate ($\rho$)} \\
    \hline
    \textbf{Cora} &  2,708 & 5,278 & 1,433 & 7 & 0.81 \\
    \textbf{Citeseer}  & 3,327 & 4,732 & 3,703 & 6 & 0.74 \\
    \textbf{Pubmed}  & 19,717 & 44,324 & 500 & 3 & 0.80 \\
    \textbf{CS}  & 18,333 & 81,894 & 6,805 & 15 & 0.81 \\
    \textbf{Physics} & 34,493 & 247,962 & 8,415 & 5 & 0.93 \\
    \midrule
    \textbf{Texas}    &  183 & 295 & 1,703 & 5 & 0.06 \\
    \textbf{Wisconsin}  & 251 & 466 & 1,703 & 5 & 0.13 \\
    \textbf{Actor}  & 7,600 & 30,019 & 932 & 5 & 0.22 \\
    \midrule
    \textbf{OGBN-} & \multirow{2}{*}{2,449,029}  & \multirow{2}{*}{61,859,140} & \multirow{2}{*}{100} & \multirow{2}{*}{47} & \multirow{2}{*}{0.81} \\
    \textbf{products} &  &  &  & &  \\
    \bottomrule
  \end{tabular}
  }
\end{table}

\section{Experimental Setup}
\subsection{Datasets}
We conduct experiments on publicly available graph datasets: three citation networks (Cora, Citeseer, Pubmed)\footnote{https://github.com/kimiyoung/planetoid/tree/master/data}, two co-author networks (CS and Physics)\footnote{https://github.com/shchur/gnn-benchmark/tree/master/data/npz}, two web page networks (Texas, Wisconsin)\footnote{https://github.com/bingzhewei/geom-gcn/tree/master/new\_data}, a movie network (Actor)\footnote{https://github.com/bingzhewei/geom-gcn/tree/master/new\_data/film}, and a large-scale co-purchasing network (OGBN-Products)\footnote{https://ogb.stanford.edu/docs/nodeprop/ogbn-products}.
Depending on whether the homophily rate exceeded 0.5, we classify homophilic graphs (Cora, Citeseer, Pubmed, CS, Physics, and OGBN-Products) and heterophilic graphs (Texas, Wisconsin, and Actor).
We provide statistics of the used datasets in Table \ref{tab1}.

\subsection{Compared Methods}
For compared methods, we employ five state-of-the-art GNN-to-MLP distillation methods: 
\begin{itemize}
    \item \textbf{GLNN} \cite{zhang2022graphless} minimizes the divergence between teacher and student outputs.

    \item \textbf{NOSMOG} \cite{tian2022nosmog} extends GLNN by considering position features,
    representational similarity distillation, and adversarial feature augmentation.
    It runs the DeepWalk algorithm on the input graph to produce the same positional features.

    \item \textbf{FF-G2M} \cite{wu2023extracting} extracts low-frequency and high-frequency knowledge from GNNs and injects it into MLPs.

    \item \textbf{HGMD} \cite{wu2024teach} distills the knowledge at the subgraph level in a hardness-aware manner.

    \item \textbf{LLP} \cite{guo2023linkless} distills relational knowledge centered on anchor nodes.
\end{itemize}

We employ 3-layer MLPs as student models in our work and all the compared methods.
The learning rate and weight decay are set to 5e-3 and 5e-4. 
For all the compared methods and DAD-SGM, the dimensions of hidden representations and final representations are set to 128 and 64, respectively.
Given the assumption that no labels are available, we assign zero weights to the classification loss of the compared methods.
For the proposed DAD-SGM, we fix the diffusion step $T$ to 20.
For NOSMOG and DAD-SGM, we use the same 64-dimensional positional encodings generated with DeepWalk using three walks per node, a 5-window size, and 20 walk lengths.
For HGMD, we use hyperparameters selected by the authors \footnote{https://github.com/LirongWu/HGMD}.
For LLP, we set the margin for rank-based distillation and the sample rate over nearby nodes as 0.1 and 1, respectively.

\begin{table*}[th]
\centering
\caption{Node classification performance on graphs with high homophily. 
\textbf{Bold} = best; \underline{Underline} = second best.}
\label{tab2}
\resizebox{0.8\textwidth}{!}{%
\begin{tabular}{c|lc|ccccc|c}
\toprule
    \textbf{Dataset} 
    & \multicolumn{2}{c|}{\textbf{Teacher GNN}}
    & \textbf{GLNN} & \textbf{NOSMOG} & \textbf{FF-G2M} & \textbf{HGMD} & \textbf{LLP}
    & \textbf{DAD-SGM (Ours)} \\
\midrule
& GAE       & 0.802 
  & 0.786(-2.00\%) & 0.788(-1.75\%) & 0.791(-1.37\%) & \underline{0.795(-0.87\%)} & 0.752(-6.23\%) 
  & \textbf{0.803(+0.12\%)} \\

& GRACE     & 0.823 
  & 0.810(-1.58\%) & 0.816(-0.85\%) & 0.816(-0.85\%) & \underline{0.824(+0.12\%)} & 0.803(-2.43\%) 
  & \textbf{0.828(+0.61\%)} \\

\textbf{Cora} & DGI       & 0.809
  & 0.774(-4.33\%) & \underline{0.789(-2.47\%)} & 0.786(-2.84\%) & 0.781(-3.46\%) & 0.783(-3.21\%)  
  & \textbf{0.813(+0.49\%)} \\

& CCA-SSG   & 0.829
  & 0.810(-2.29\%) & 0.822(-0.84\%) & 0.806(-2.77\%) & \underline{0.825(-0.48\%)} & 0.791(-4.58\%)  
  & \textbf{0.829(+0.00\%)} \\

& GBT       & 0.831
  & 0.808(-2.77\%) & \underline{0.820(-1.32\%)} & 0.813(-2.17\%) & 0.816(-1.81\%) & 0.782(-5.90\%)  
  & \textbf{0.832(+0.12\%)} \\

\midrule
& GAE       & 0.688 
  & 0.661(-3.92\%) & 0.681(-1.02\%) & 0.681(-1.02\%) & \underline{0.691(+0.44\%)} & 0.625(-9.16\%)
  & \textbf{0.712(+3.49\%)} \\

& GRACE     & 0.714
  & 0.703(-1.54\%) & 0.708(-0.84\%) & 0.706(-1.12\%) & \underline{0.714(+0.00\%)} & 0.709(-0.70\%)  
  & \textbf{0.715(+0.14\%)} \\

\textbf{Citeseer} & DGI       & 0.717
  & 0.683(-4.74\%) & 0.694(-3.21\%) & 0.701(-2.23\%) & \underline{0.709(-1.12\%)} & 0.666(-7.11\%) 
  & \textbf{0.713(-0.56\%)} \\

& CCA-SSG   & 0.708
  & 0.699(-1.27\%) & \underline{0.703(-0.71\%)} & 0.692(-2.26\%) & \underline{0.703(-0.71\%)} & 0.701(-0.99\%) 
  & \textbf{0.711(+0.42\%)} \\

& GBT       & 0.718
  & 0.710(-1.11\%) & 0.705(-1.81\%) & 0.701(-2.36\%) & \underline{0.716(-0.28\%)} & 0.704(-1.95\%) 
  & \textbf{0.724(+0.84\%)} \\

\midrule
& GAE       & 0.789
  & 0.767(-2.79\%) & 0.770(-2.41\%) & 0.785(-0.51\%) & \underline{0.791(+0.25\%)} & 0.765(-3.04\%)
  & \textbf{0.803(+1.77\%)} \\

& GRACE     & 0.793
  & 0.780(-1.64\%) & \underline{0.791(-0.25\%)} & 0.790(-0.38\%) & \underline{0.797(+0.50\%)} & 0.787(-0.76\%) 
  & \textbf{0.801(+1.01\%)} \\

\textbf{Pubmed} & DGI       & 0.802
  & 0.785(-2.12\%) & 0.772(-3.74\%) & 0.788(-1.75\%) & \underline{0.799(-0.37\%)} & 0.729(-9.10\%) 
  & \textbf{0.819(+2.12\%)} \\

& CCA-SSG   & 0.799
  & 0.751(-6.01\%) & \underline{0.783(-2.00\%)} & 0.781(-2.25\%) & 0.774(-3.13\%) & 0.681(-14.77\%) 
  & \textbf{0.804(+0.63\%)} \\

& GBT       & 0.789
  & \underline{0.764(-3.17\%)} & 0.755(-4.31\%) & 0.751(-4.82\%) & 0.717(-9.13\%) & 0.705(-10.65\%) 
  & \textbf{0.794(+0.63\%)} \\

\midrule
& GAE       & 0.918
  & 0.930(+1.32\%) & 0.933(+1.65\%) & 0.936(+1.98\%) & \underline{0.940(+2.41\%)} & 0.928(-1.10\%)
  & \textbf{0.943(+2.73\%)} \\

& GRACE     & 0.925
  & \underline{0.928(+0.32\%)} & 0.927(+0.22\%) & 0.921(-0.43\%) & 0.919(-0.65\%) & 0.927(+0.22\%) 
  & \textbf{0.933(+0.86\%)} \\

\textbf{CS} & DGI       & 0.922
  & \underline{0.926(+0.41\%)} & 0.896(-2.82\%) & 0.901(-2.28\%) & 0.909(-1.41\%) & 0.914(-0.87\%)  
  & \textbf{0.929(+0.76\%)} \\

& CCA-SSG   & 0.918
  & \underline{0.911(-0.76\%)} & 0.902(-1.74\%) & 0.886(-3.49\%) & 0.889(-3.16\%) & 0.907(-1.20\%)  
  & \textbf{0.921(+0.33\%)} \\

& GBT       & 0.922
  & \underline{0.913(-0.98\%)} & 0.909(-1.43\%) & 0.881(-4.45\%) & 0.892(-3.25\%) & 0.908(-1.52\%)  
  & \textbf{0.930(+0.92\%)} \\

\midrule
& GAE       & 0.948
  & 0.944(-0.42\%) & 0.941(-0.74\%) & 0.941(-0.74\%) & \textbf{0.953(+0.53\%)} & 0.943(-0.53\%)
  & \textbf{0.953(+0.53\%)} \\

& GRACE     & 0.953
  & 0.952(-0.10\%) & \underline{0.953(+0.00\%)} & 0.952(-0.10\%) & 0.950(-0.31\%) & 0.952(-0.10\%)  
  & \textbf{0.956(+0.27\%)} \\

\textbf{Physics} & DGI       & 0.952
  & \underline{0.945(-0.76\%)} & 0.944(-0.82\%) & 0.935(-1.78\%) & 0.940(-1.24\%) & 0.939(-1.34\%) 
  & \textbf{0.954(+0.20\%)} \\

& CCA-SSG   & 0.953
  & \underline{0.947(-0.59\%)} & 0.945(-0.88\%) & 0.940(-1.39\%) & 0.944(-0.94\%) & 0.945(-0.84\%) 
  & \textbf{0.953(+0.00\%)} \\

& GBT       & 0.952
  & \underline{0.950(-0.24\%)} & 0.948(-0.39\%) & 0.942(-1.05\%) & 0.944(-0.83\%) & 0.941(-1.15\%) 
  & \textbf{0.952(+0.00\%)} \\

\bottomrule
\end{tabular}
}
\end{table*}

\begin{table*}[th]
    \caption{Node classification performance on graphs with low homophily.
    \textbf{Bold} = best; \underline{Underline} = second best.}
    \label{tab3}
    \centering
    \resizebox{0.8\textwidth}{!}{%
    \begin{tabular}{c|lc|ccccc|c}
    \toprule
    \textbf{Dataset} 
    & \multicolumn{2}{c|}{\textbf{Teacher GNN}}
    & \textbf{GLNN} & \textbf{NOSMOG} & \textbf{FF-G2M} & \textbf{HGMD} & \textbf{LLP}
    & \textbf{DAD-SGM (Ours)} \\

\midrule
\multirow{2}{*}{\textbf{Texas}} 
  & SELENE       & 0.730
  & 0.730(+0.00\%) & 0.703(-3.70\%) & \textbf{0.757(+3.70\%)} & \textbf{0.757(+3.70\%)} & 0.730(+0.00\%)  & \textbf{0.757(+3.70\%)} \\
  & HGRL       & 0.730
  & 0.730(+0.00\%) & 0.703(-3.70\%) & 0.730(+0.00\%) & 0.649(-11.15\%) & \textbf{0.757(+3.70\%)}  & \textbf{0.757(+3.70\%)} \\

\midrule
\multirow{2}{*}{\textbf{Wisconsin}}
  & SELENE       & 0.745
  & \underline{0.745(+0.00\%)} & 0.725(-2.63\%) & 0.647(-13.15\%) & 0.725(-2.63\%) & \underline{0.745(+0.00\%)} 
  & \textbf{0.784(+5.23\%)} \\

  & HGRL       & 0.804
  & 0.676(-15.92\%) & \textbf{0.804(+0.00\%)} & 0.686(-14.65\%) & 0.745(-7.34\%) & 0.745(-7.34\%)
  & \textbf{0.804(+0.00\%)} \\
  
\midrule
\multirow{2}{*}{\textbf{Actor}}
& SELENE   & 0.365
  & 0.365(+0.00\%) & \underline{0.368(+0.82\%)} & 0.360(-1.37\%) & 0.349(-4.49\%) & 0.365(+0.00\%)
  & \textbf{0.371(+1.64\%)} \\

& HGRL       & 0.359
  & 0.357(-0.36\%) & \textbf{0.368(+2.62\%)} & 0.367(+2.23\%) & 0.304(-15.23\%) & 0.364(+1.53\%)
  & \textbf{0.368(+2.62\%)} \\
    
    \bottomrule
    \end{tabular}
    }
\end{table*}

\begin{table*}[th]
\centering
\caption{Link prediction performance on graphs with high homophily. 
\textbf{Bold} = best; \underline{Underline} = second best.}
\label{tab4}
\resizebox{0.8\textwidth}{!}{%
\begin{tabular}{c|lc|ccccc|c}
\toprule
    \textbf{Dataset} 
    & \multicolumn{2}{c|}{\textbf{Teacher GNN}}
    & \textbf{GLNN} & \textbf{NOSMOG} & \textbf{FF-G2M} & \textbf{HGMD} & \textbf{LLP}
    & \textbf{DAD-SGM (Ours)} \\
\midrule
& GAE
  & 0.922 
  & 0.927(+0.54\%) & 0.917(-0.54\%) & 0.913(-0.98\%) & 0.903(-2.06\%) & \underline{0.929(+0.76\%)} & \textbf{0.934(+1.30\%)} \\

& GRACE
  & 0.947 
  & 0.939(-0.79\%) & 0.932(-1.58\%) & 0.923(-2.53\%) & 0.845(-10.77\%) & \underline{0.944(-0.32\%)} & \textbf{0.954(+0.72\%)} \\

\textbf{Cora} & DGI
  & 0.916
      & \underline{0.913(-0.33\%)} & 0.904(-1.31\%) & 0.880(-3.93\%) & 0.912(-0.44\%) & 0.898(-1.97\%) & \textbf{0.965(+5.41\%)} \\

& CCA-SSG
  & 0.936  & 0.925(-1.18\%) & 0.921(-1.60\%) & 0.908(-2.99\%) & 0.797(-14.85\%) & \underline{0.931(-0.53\%)} & \textbf{0.949(+1.39\%)} \\

& GBT
  & 0.931
  & 0.924(-0.75\%) & 0.906(-2.69\%) & 0.903(-3.01\%) & 0.919(-1.31\%) & \underline{0.927(-0.43\%)} & \textbf{0.944(+1.43\%)} \\
\midrule
& GAE
  & 0.946
  & 0.913(-3.49\%) & 0.923(-2.43\%) & 0.875(-7.51\%) & 0.902(-4.65\%) & \underline{0.937(-0.95\%)} & \textbf{0.944(-0.21\%)} \\

& GRACE
  & 0.957
  & \underline{0.959(+0.21\%)} & 0.956(-0.10\%) & 0.944(-1.36\%) & 0.922(-3.66\%) & 0.956(-0.10\%) & \textbf{0.960(+0.31\%)} \\

\textbf{Citeseer} & DGI
  & 0.923
  & 0.919(-0.43\%) & 0.917(-0.65\%) & 0.866(-6.18\%) & 0.930(+0.76\%) & \underline{0.933(+1.08\%)} & \textbf{0.959(+3.94\%)} \\

& CCA-SSG
  & 0.948
  & 0.947(-0.11\%) & 0.945(-0.32\%) & 0.933(-1.58\%) & 0.885(-6.65\%) & \underline{0.949(+0.11\%)} & \textbf{0.957(+0.95\%)} \\

& GBT
  & 0.948
  & \underline{0.948(+0.00\%)} & 0.941(-0.74\%) & 0.933(-1.58\%) & 0.931(+1.79\%) & \underline{0.948(+0.00\%)} & \textbf{0.953(+0.53\%)} \\
\midrule
& GAE
  & 0.946
  & 0.923(-2.43\%) & 0.914(-3.38\%) & 0.907(-4.12\%) & 0.912(-3.59\%) & \textbf{0.942(-0.42\%)} & \underline{0.940(-0.63\%)} \\

& GRACE
  & 0.934
  & 0.899(-3.75\%) & 0.885(-5.25\%) & 0.886(-5.14\%) & 0.888(-4.93\%) & \textbf{0.939(+0.54\%)} & \underline{0.934(+0.00\%)} \\

\textbf{Pubmed} & DGI
  & 0.889
  & 0.859(-3.37\%) & 0.823(-7.42\%) & 0.816(-8.21\%) & 0.842(-5.29\%) & \underline{0.909(+2.25\%)} & \textbf{0.938(+5.51\%)} \\

& CCA-SSG
  & 0.942
  & 0.895(-4.99\%) & 0.865(-8.17\%) & 0.852(-9.55\%) & 0.861(-8.60\%) & \underline{0.921(-2.23\%)} & \textbf{0.929(-1.38\%)} \\

& GBT
  & 0.941
  & 0.883(-6.16\%) & 0.849(-9.78\%) & 0.848(-9.88\%) & 0.859(-8.71\%) & \underline{0.919(-2.34\%)} & \textbf{0.929(-1.28\%)} \\
\midrule
& GAE
  & 0.947
  & 0.951(+0.42\%) & 0.931(-1.69\%) & 0.943(-0.42\%) & 0.917(-3.13\%) & \textbf{0.959(+1.27\%)} & \textbf{0.959(+1.27\%)} \\

& GRACE
  & 0.943
  & 0.941(-0.21\%) & 0.937(-0.64\%) & 0.931(-1.27\%) & 0.943(+0.00\%) & \underline{0.946(+0.32\%)} & \textbf{0.947(+0.42\%)} \\

\textbf{CS} & DGI
  & 0.942
  & \underline{0.937(-0.53\%)} & 0.891(-5.41\%) & 0.879(-6.69\%) & 0.915(-2.87\%) & 0.932(-1.06\%) & \textbf{0.948(+0.64\%)} \\

& CCA-SSG
  & 0.936
  & 0.933(-0.32\%) & 0.924(-1.28\%) & 0.932(-0.43\%) & 0.933(-0.32\%) & \underline{0.935(-0.11\%)} & \textbf{0.942(+0.64\%)} \\

& GBT
  & 0.935
  & 0.932(-0.32\%) & 0.917(-1.93\%) & 0.915(-2.14\%) & 0.933(-0.21\%) & \underline{0.937(+0.21\%)} & \textbf{0.942(+0.75\%)} \\
\midrule
& GAE
  & 0.952
  & 0.944(-0.84\%) & 0.918(-3.57\%) & 0.943(-0.95\%) & 0.926(-2.73\%) & \textbf{0.960(+0.84\%)} & \textbf{0.960(+0.84\%)} \\

& GRACE
  & 0.949
  & 0.944(-0.52\%) & 0.942(-0.73\%) & 0.933(-1.68\%) & 0.911(-0.62\%) & \underline{0.951(+0.22\%)} & \textbf{0.955(+0.64\%)} \\

\textbf{Physics} & DGI
  & 0.951
  & 0.935(-1.68\%) & 0.914(-3.89\%) & 0.883(-7.15\%) & 0.938(-1.37\%) & \underline{0.946(-0.53\%)} & \textbf{0.953(+0.21\%)} \\

& CCA-SSG
  & 0.942
  & \underline{0.933(-0.96\%)} & 0.919(-2.44\%) & 0.883(-6.26\%) & 0.906(-3.82\%) & 0.930(-1.13\%) & \textbf{0.944(+0.21\%)} \\

& GBT
  & 0.943
  & 0.925(-1.91\%) & 0.921(-2.33\%) & 0.905(-4.03\%) & 0.926(-1.80\%) & \underline{0.932(-1.17\%)} & \textbf{0.943(+0.00\%)} \\
\bottomrule
\end{tabular}
}
\end{table*}

\begin{table*}[th]
    \caption{Link prediction performance on graphs with low homophily.
    \textbf{Bold} = best; \underline{Underline} = second best.}
    \label{tab5}
    \centering
    \resizebox{0.8\textwidth}{!}{%
    \begin{tabular}{c|lc|ccccc|c}
    \toprule
        \textbf{Dataset} 
        & \multicolumn{2}{c|}{\textbf{Teacher GNN}}
        & \textbf{GLNN} & \textbf{NOSMOG} & \textbf{FF-G2M} & \textbf{HGMD} & \textbf{LLP}
        & \textbf{DAD-SGM (Ours)} \\

\midrule
\multirow{2}{*}{\textbf{Texas}} 
  & SELENE       & 0.730
  & 0.679(-0.88\%) & 0.675(-1.43\%) & 0.679(-0.88\%) & 0.596(-18.36\%) & \textbf{0.757(+3.70\%)} & \textbf{0.757(+3.70\%)} \\
  & HGRL       & 0.703
  & 0.633(-9.96\%) & 0.662(-5.83\%) & 0.668(-4.98\%) & 0.563(-19.99\%) & \textbf{0.746(+6.12\%)} & \textbf{0.746(+6.12\%)} \\

\midrule
\multirow{2}{*}{\textbf{Wisconsin}}
  & SELENE       & 0.652
  & 0.620(-4.91\%) & 0.679(+4.14\%) & 0.652(+0.00\%) & \textbf{0.729(+11.90\%)} & 0.679(+4.14\%)
  & \underline{0.704(+8.13\%)} \\

  & HGRL       & 0.779
  & \underline{0.802(+2.91\%)} & 0.775(-0.49\%) & \underline{0.802(+2.91\%)} & 0.734(-5.75\%) & 0.774(-0.62\%)
  & \textbf{0.817(+4.90\%)} \\
  
\midrule
\multirow{2}{*}{\textbf{Actor}}
& SELENE   & 0.516
  & 0.506(-1.96\%) & 0.508(-1.55\%) & 0.509(-1.36\%) & 0.507(-1.76\%) & \textbf{0.527(+2.13\%)}
  & \underline{0.523(+1.36\%)} \\

& HGRL       & 0.524
  & 0.528(+0.76\%) & 0.521(-0.57\%) & 0.530(+1.15\%) & 0.511(-2.48\%) & \underline{0.530(+1.15\%)}
  & \textbf{0.533(+1.72\%)} \\
    
    \bottomrule
    \end{tabular}
    }
\end{table*}

\begin{table}[t]
\centering
\caption{Node Classification performance with a \textbf{supervised} GCN teacher.
\textbf{Bold} = best; \underline{Underline} = second best.
}
\label{tab:sup_teacher}
\resizebox{\linewidth}{!}{%
\begin{tabular}{lc|ccccc}
\toprule
 & \textbf{GCN} & \multirow{2}{*}{\textbf{GLNN}} & \multirow{2}{*}{\textbf{NOSMOG}} & \multirow{2}{*}{\textbf{FF-G2M}} & \multirow{2}{*}{\textbf{HGMD}} & \textbf{DAD‑SGM} \\
 & \textbf{(Teacher)} &  &  &  &  & \textbf{(Ours)} \\
\midrule
\textit{Use Label?} & \ding{51} & \ding{51} & \ding{51} & \ding{51} & \ding{51} & \ding{55} \\
\textit{Used Info.} &  & logits & logits + repr & logits & logits & repr \\
\midrule
Cora        & 0.818 & 0.822 & 0.825 & 0.829 & \underline{0.831} & \textbf{0.836} \\
Citeseer    & 0.702 & 0.724 & 0.717 & \underline{0.737} & 0.734 & \textbf{0.739} \\
Pubmed      & 0.799 & 0.794 & 0.805 & \underline{0.808} & 0.791 & \textbf{0.814} \\
CS          & 0.932 & 0.935 & 0.938 & \underline{0.944} & 0.939 & \textbf{0.946} \\
Physics     & 0.956 & 0.957 & 0.956 & 0.956 & \textbf{0.959} & \textbf{0.959} \\
\bottomrule
\end{tabular}}
\end{table}

\subsection{Teacher Models}
Our DAD-SGM aims to distill self-supervised GNNs into MLPs effectively. 
We select a range of self-supervised GNNs as teacher models to assess how well our work accomplishes this. 
Specifically, we employ GAE \cite{kipf2016variational}, GRACE \cite{zhu2020deep}, DGI \cite{veličković2018deep}, CCA-SSG \cite{zhang2021canonical}, and GBT \cite{bielak2022graph} for graphs with high-homophily.
For low-homophily graphs, we use SELENE \cite{zhong2022unsupervised} and HGRL \cite{chen2022towards}.

Across all these teacher models, we set the number of layers and the dimensions of node representations as 3 and 64, respectively.
For all teacher models, we set the dimensions of hidden representations and final representations to 256 and 64, respectively.
GAE utilizes negative samples of the same size as the positive edges.
DGI derives corrupted features by shuffling node features in a row-wise manner.
We utilize the mean-pooling to compute its summary vector.
For all methods, we employ hyperparameters selected by the authors.

\section{Results \& Analysis}
\subsection{Node Classification}
We conduct a logistic regression on node representations five times and compute the average Micro-F1 scores.
Micro F1 score is less sensitive to data imbalance issues and accurately reflects overall classification performance.
The score is calculated based on the True Positives, False Positives, and False Negatives.
Table \ref{tab2} and Table \ref{tab3} show the node classification performance of DAD-SGM and the compared methods on homophilic and heterophilic graphs, respectively.
The tables also show the performance improvement rates of student models relative to their pre-trained teachers.

From these tables, we observe that the effectiveness of GNN-to-MLP distillation methods varies depending on the teacher's underlying strategy. 
For instance, NOSMOG and FF-G2M can effectively distill GRACE, which learns representations based on node similarities.
This may be because NOSMOG and FF-G2M focus on node similarity information during the distillation process.
However, they struggle to capture the decorrelation-based representations learned by CCA-SSG and GBT.
We also observe that LLP, designed for link prediction, can reduce node classification performance by up to 15\%.
In contrast, DAD-SGM consistently achieves the best score in distilling diverse self-supervised GNNs. 
We also evaluate DAD-SGM in inductive node classification, and the results are provided in Appendix \ref{appendix}.


\subsection{Link Prediction}  
We also evaluate the link prediction capability of DAD-SGM using the \emph{Area Under the ROC Curve} (AUC-ROC) metric, which assesses a model's ability to predict the existence of links in graphs.
A higher AUC-ROC score indicates better predictive performance, with a score closer to 1 implying near-perfect prediction.
We train teacher GNNs and positional encodings for each graph using 85\% of the edges.
The remaining edges are divided into a validation set and a test set at a ratio of 1 to 2.
We add the same number of randomly sampled negative edges to the validation and test sets.

Tables~\ref{tab4} and \ref{tab5} show that node representations of NOSMOG, FF-G2M, and HGMD fail to maintain the teachers' link prediction performance.
In contrast, node representations from LLP show strong results in link prediction but are less effective at node classification.
By contrast, DAD-SGM excels at both tasks, surpassing state-of-the-art baselines and teacher models in most cases.
These findings demonstrate that DAD-SGM can produce high-quality node representations that perform well across diverse tasks.

\subsection{Distilling a Supervised Teacher}
The proposed DAD-SGM is designed for distilling self-supervised GNNs, so it utilizes node representations of the teacher models.
However, practical scenarios often require distilling supervised GNNs.
To examine whether DAD-SGM remains effective in such cases, we compare our method against other GNN-to-MLP distillation approaches in distilling a 3-layer GCN trained in a supervised manner.
Table~\ref{tab:sup_teacher} presents the results.  
Here, DAD-SGM's classification performance is computed via logistic regression on its learned node representations.
Despite not relying on labels or teacher logits, DAD-SGM matches or exceeds the performance of existing GNN-to-MLP distillation methods. 
This demonstrates that our approach retains its effectiveness even when the teacher GNN is trained under supervision.
{\color{black}{Appendix C shows the results other semi-supervised teachers: GAT \cite{veličković2018graph} and APPNP \cite{gasteiger2018combining}.}}

\begin{figure}[t]
    \centering
    \centerline{\includegraphics[width=0.48\textwidth]{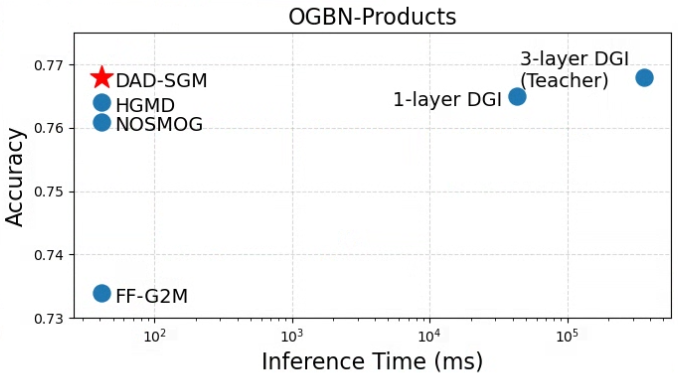}}
    \caption{Inference time (ms) and accuracy (F1 score) on the Ogbn-products dataset.}
    \label{fig4}
\end{figure}

\subsection{Large-Scale Dataset}
To investigate the effectiveness of our work in real-world scenarios, we compare the classification accuracy and inference speed of our work and other state-of-the-art GNN-to-MLP distillation methods (NOSMOG, FF-G2M, and HGMD) on a large-scale OGBN-Products.
The dataset consists of 2 million nodes and 61 million edges.
Figure \ref{fig4} shows that our work and other GNN-to-MLP distillation methods drastically decrease the inference time compared to GNNs.
They are 1,048 times and 8,781 times faster than 1-layer and 3-layer DGI models, respectively.
Meanwhile, our work achieves the best score among GNN-to-MLP distillation methods.
These results highlight the practical potential of DAD-SGM for large-scale applications, where both accuracy and efficiency are crucial considerations.

\subsection{Ablation Study}
We propose an MLP-architected denoising diffusion model as a teacher assistant (TA) to narrow the large capacity gap between GNNs and MLPs in SSGRL.
This section presents two ablation studies to quantify the effectiveness of the proposed TA.
First, we compare our TA against several alternative TA models.
Second, we benchmark our TA against data augmentation techniques that can also mitigate the teacher-student gap in knowledge distillation.

\subsubsection{Comparison with Other TA Candidates}
Our TA leverages a denoising diffusion process to facilitate the smooth knowledge transfer between self-supervised GNNs and MLPs.
However, it may also be possible to employ other approaches, such as contrastive learning or masked autoencoding.
To assess the effectiveness of our diffusion-based TA, we compare it against four alternative TAs that are distilled from the same GRACE teacher with distinct strategies:
\begin{itemize}
    \item 
    FitNet \cite{DBLP:journals/corr/RomeroBKCGB14} minimizes the mean squared error (MSE) between teacher and student representations.
    \item
    Similarity Preserving (SP) \cite{tung2019similarity} aligns the similarity maps derived from teacher and student representations.
    \item
    Contrastive Representation Distillation (CRD) \cite{tian2019contrastive} distills the representational knowledge into the student via contrastive learning.
    \item
    Masked Knowledge Distillation (MKD) \cite{lao2023masked} employs a masked autoencoding scheme to enhance the knowledge distillation.
\end{itemize}
Each baseline is implemented with both a 1‑layer GCN and a 3‑layer MLP.
We also include a variant of our own TA that replaces the noise-prediction network $\epsilon_\phi$ with a 1-layer GCN.

\begin{figure}
    \centering
    \centerline{\includegraphics[width=0.53\textwidth]{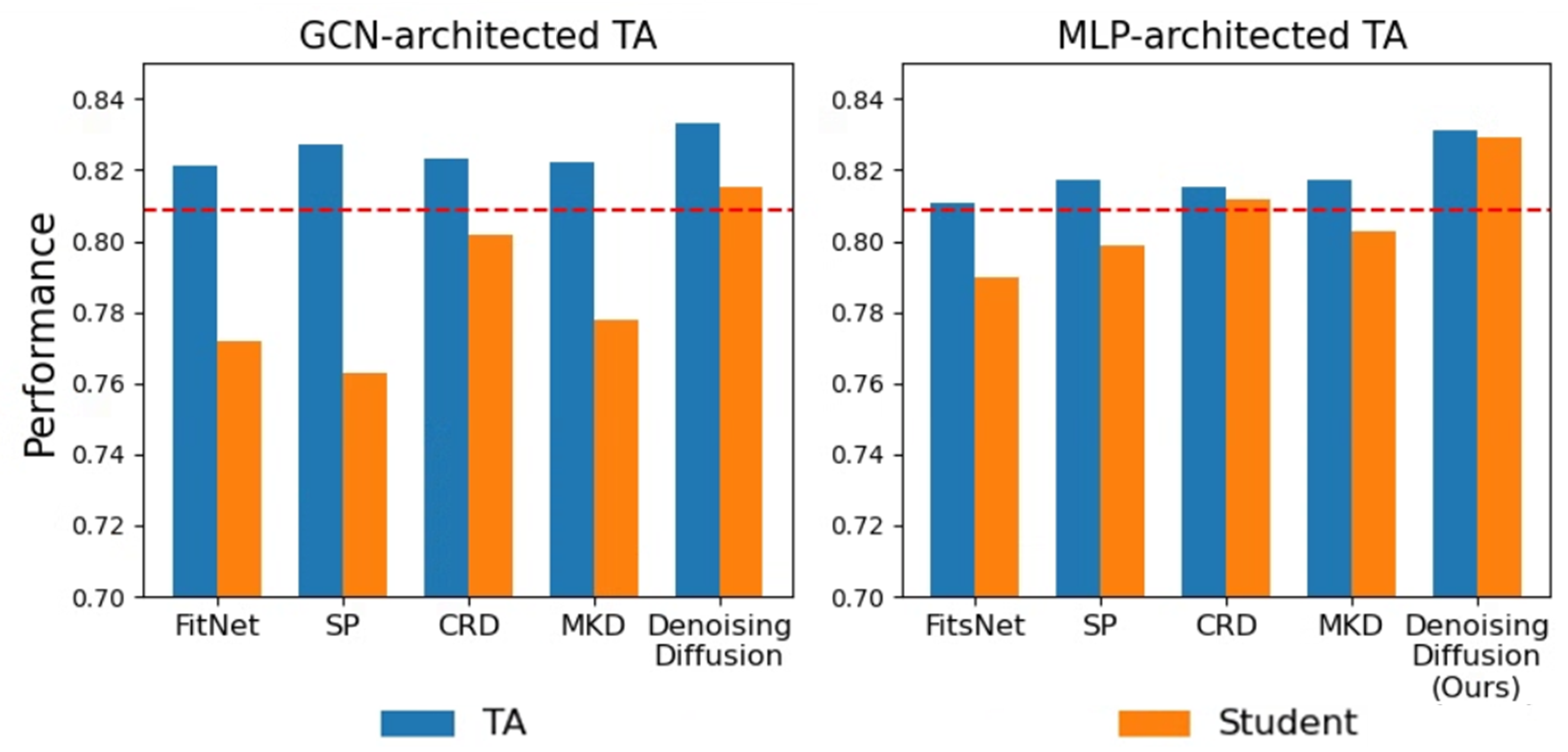}}
    \vspace{-1em}
    \caption{Comparison of our TA model with other candidates. 
    Blue bars represent the node classification performance of TAs, and orange bars represent that of their distilled students. 
    The red dotted line shows the performance of a baseline student model directly distilled from the teacher (without TA).}
    \label{fig5}
    \vspace{-0.5em}
\end{figure}

Figure \ref{fig5} presents the node classification accuracy on Cora with a GRACE teacher. 
Blue bars represent the accuracy of the TAs, while orange bars indicate the accuracy of the student MLPs.
The red dotted line shows the performance of a baseline student model directly distilled from the teacher (i.e., no TA).
From this figure, we can see that MLP-architected TAs capture the teachers' knowledge slightly less accurately than GCN-architected TAs.
However, their students achieve higher accuracy than those trained with GCN-based TAs.
This is likely due to architectural compatibility with the MLP students and MLP-architected TAs.
At the same time, diffusion-based TAs are beneficial in improving student performance across both architectures, unlike other TA candidates.
These observations support that our TA model is more effective at reducing the teacher–student capacity gap in SSGRL, outperforming contrastive and masked autoencoding approaches.

\begin{figure}
    \centering
    \centerline{\includegraphics[height=5.3cm]{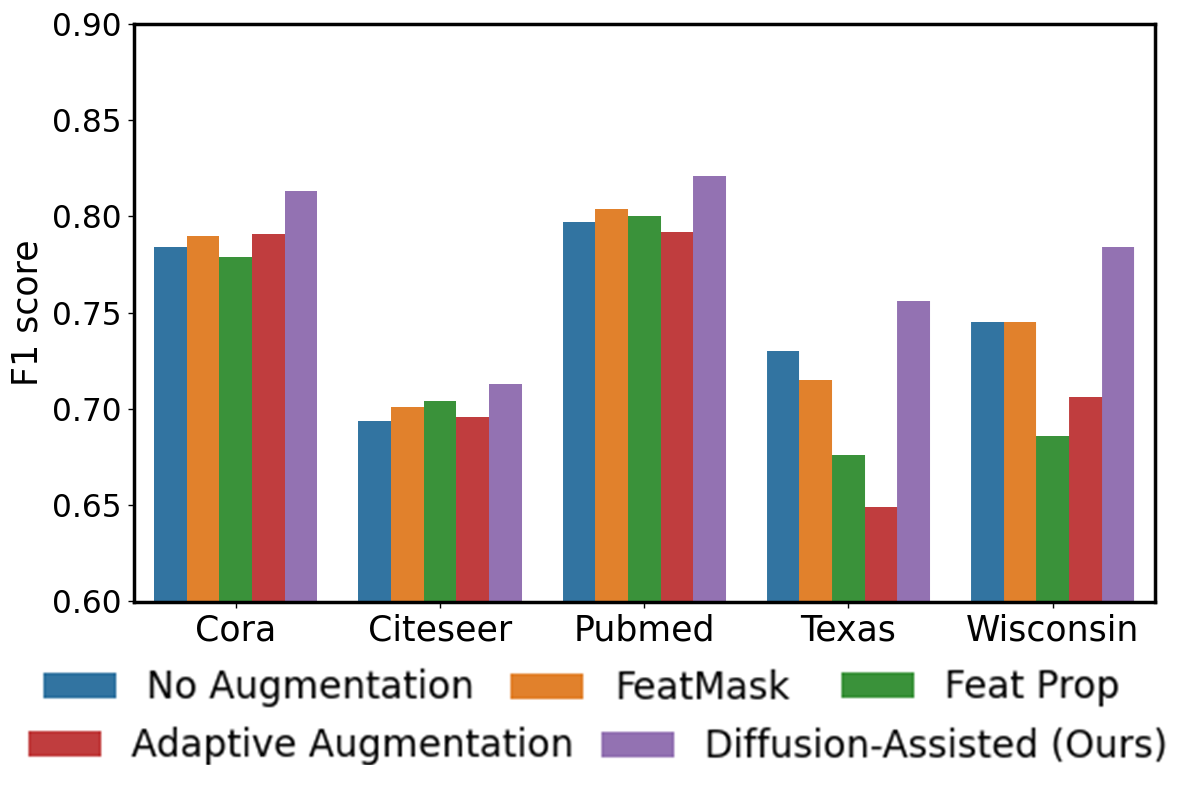}}
    \vspace{-1em}
    \caption{Comparison with graph data augmentation techniques.}
    \label{fig6}
    \vspace{-1.5em}
\end{figure}

\subsubsection{Comparison with Graph Data Augmentation}
Data augmentation also serves as a strategy to bridge the capacity gap between teacher and student models \cite{das2020empirical,wang2023understanding,li2022role}.
Hence, we also compare our diffusion-based TA with three representative graph‑augmentation schemes: FeatMask \cite{zhao2021data}, Feature Propagation \cite{feng2020graph}, and Adaptive Augmentation \cite{zhu2021graph}.
We implement three variants of GLNN that incorporate these augmentation strategies: GLNN$_\text{FeatMask}$, GLNN$_\text{FeatProp}$, and GLNN$_\text{AdapAug}$.
We tune hyperparameters of the variants via grid search, considering feature drop rates ($\alpha\in\{0.1, 0.2, 0.3, 0.4,$ $0.5\}$) and edge drop rates ($\beta\in\{0.05, 0.1, 0.15, 0.2, 0.25, 0.3\}$).
For a fair comparison, we train the variants with the concatenation of node features and positional encodings.
For three homophilic graphs (Cora, Citeseer, and Pubmed) and two heterophilic graphs (Texas and Wisconsin), we measure F1 scores to investigate the impact of our method and other graph data augmentation strategies on GNN-to-MLP distillation.
Figure \ref{fig6} shows that our diffusion-based TA is the only method that consistently surpasses vanilla GLNN across all datasets.
In contrast, other GLNN variants often have lower F1 scores than GLNN, especially in heterophilic graphs.
We believe this aligns with prior findings that data augmentation can disrupt essential characteristics of graph-structured data (e.g., dropping nodes/edges may disrupt chemical bonds or functional groups) \cite{lee2022augmentation,xia2022simgrace}.
This demonstrates the benefit of our diffusion-based TA in enhancing the power of MLPs across graphs with diverse characteristics.

\subsection{Noise Robustness}
\begin{figure}[t]      
    \centering
    \centerline{\includegraphics[width=0.48\textwidth]{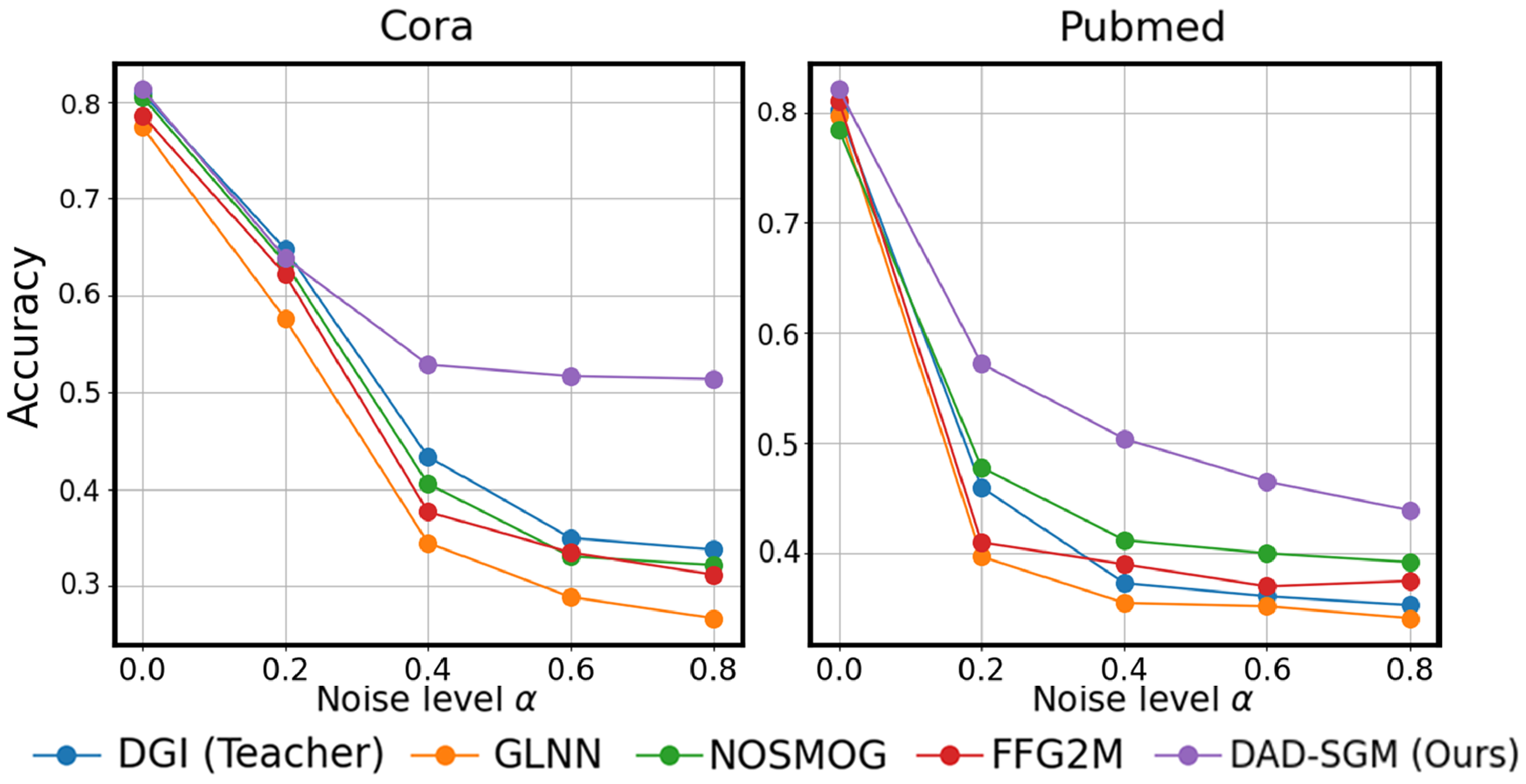}}
    \caption{Node classification performance (accuracy) with different levels of Gaussian noise.}
    \label{fig7}

    \centerline{\includegraphics[width=0.48\textwidth]{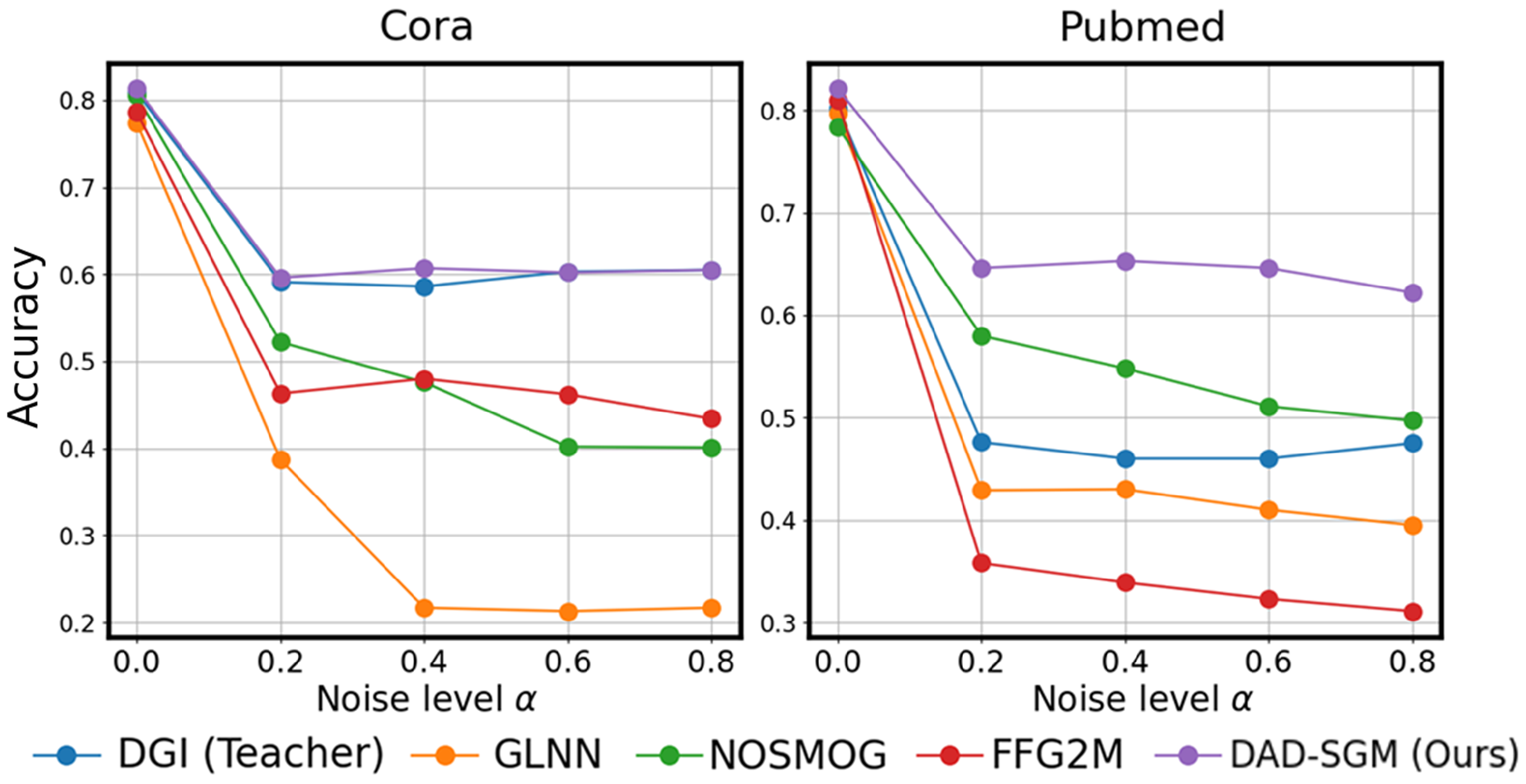}}
    \caption{Node classification performance (accuracy) with different levels of FGSM noise.}
    \label{fig8}
\end{figure}

A well-known drawback of MLPs in graph analysis is their sensitivity against feature noises \cite{zhang2022graphless,tian2022nosmog}.
We conduct two experiments to investigate whether DAD-SGM enhances the robustness of MLPs in self-supervised graph representation learning. 
Firstly, we perturb node features with Gaussian noise $\epsilon\sim \mathcal{N}{(0,\mathbf{I})}$ by replacing $\textit{\textbf{x}}_v$ with $(1-\alpha)\textit{\textbf{x}}_v+\alpha\cdot \epsilon$ for each node $v$,  where $\alpha$ controls the noise level.
Figure \ref{fig7} illustrates the F1 scores achieved by our methods and other GNN-to-MLP distillation methods on the Cora and Pubmed datasets.
As the noise level $\alpha$ increases, DAD-SGM not only outperforms the compared methods but also surpasses the performance of the teacher model.
This observation suggests that our approach significantly enhances the robustness of MLP students against Gaussian noise.

Also, we investigate the robustness of DAD-SGM against Fast Gradient Sign Method (FGSM) attacks \cite{goodfellow2014explaining}.
FGSM is one of the representative techniques that adds subtle noise that deceives deep learning models.
We produce the adversarial example of $v$ with FGSM as $x_v^*=x_v+\alpha\cdot sgn(\nabla_x\mathcal{L}(h^{\text{Stu}}_v,h^{\text{Tea}}_v))$, where $\mathcal{L}$ is a loss function of each model and $\alpha$ is the degree of noise.
Figure \ref{fig8} displays F1 scores of our methods and other GNN-to-MLP distillation methods in Cora and Pubmed.
The figures show that DAD-SGM is also robust against adversarial examples, different from existing GNN-to-MLP distillation methods.

\subsection{Hyperparameter Analysis}
\begin{figure}
    \centering
    \centerline{\includegraphics[width=0.5\textwidth]{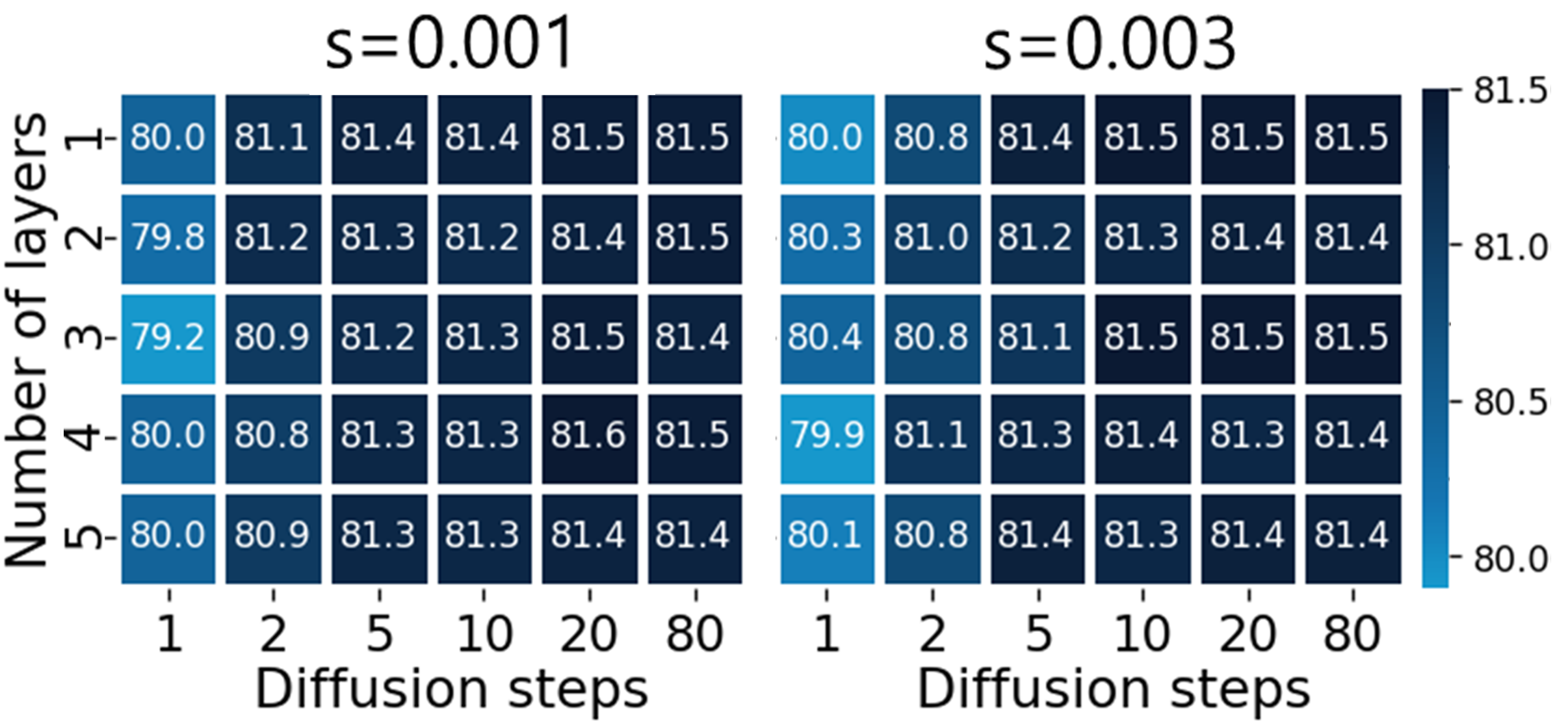}}
    \caption{Performance of DAD-SGM \emph{w.r.t.} an offset ($s$), MLP layers, and diffusion steps ($T$).}
    \label{fig9}
\end{figure}

We conduct an empirical analysis to understand the impact of three hyperparameters: the offset $s$, the number of MLP layers, and the number of diffusion steps. 
The offset $s$ controls the rate at which noise is added in the forward diffusion process. 
A larger $s$ leads to more rapid noise injection, resulting in smoother transitions between noisy samples.
We assess the F1 scores of DAD-SGM on the Cora dataset using various hyperparameter combinations as Figure \ref{fig9}. 

Our results indicate that DAD-SGM's performance is robust to the offset $s$, implying that the specific noise scheduling scheme has a relatively little impact on the overall performance.
Note that the performance of DAD-SGM is poor when $T=1$.
This shows that DAD-SGM does not perform well when the influence of the diffusion process is minimal. 
In contrast, the performance of DAD-SGM improves as $T$ increases. 
The observations show that the diffusion-based assistant model contributes to performance enhancement.
However, as the diffusion steps $T>10$, the performance of DAD-SGM converges regardless of the number of MLP layers.
This suggests that our assistant model requires small diffusion steps to produce powerful node representations.

\begin{figure}[t]
    \centerline{\includegraphics[width=0.49\textwidth]{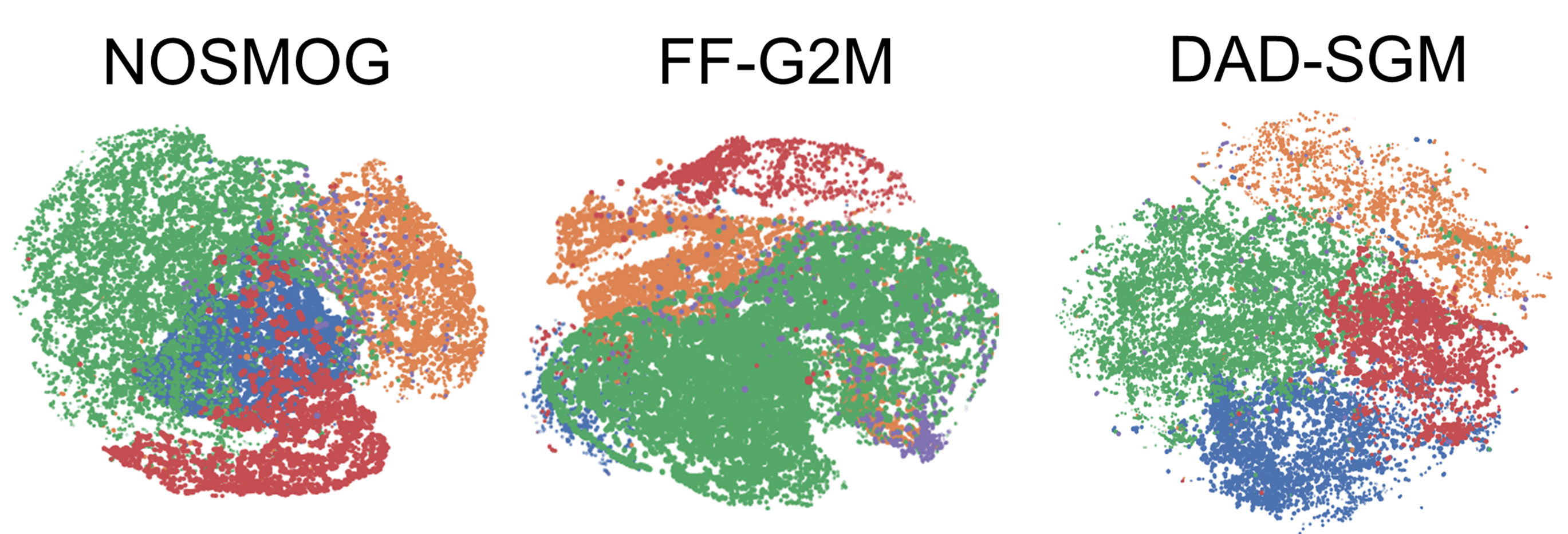}}
    \caption{Visualization of node representations in Physics.
    Each point and its color indicate an author and his / her most active field, respectively.}
    \label{fig10}
    \vspace{-1em}
\end{figure}

\subsection{Visualization}
We visually show node representations of DAD-SGM to provide a more intuitive comparison.
Figure~\ref{fig10} shows the t-SNE projection of node representations produced by three GNN-to-MLP distillation methods (NOSMOG, FF-G2M, and DAD-SGM) on the Physics dataset, each using a 3-layer DGI teacher.
The figure indicates that node representations of NOSMOG and FF-G2M are not well clustered according to their classes.
In contrast, the node representations generated by DAD-SGM are better clustered by their classes.
These findings indicate the power of our DAD-SGM in producing powerful node representations.

\section{Conclusion}
This paper presents a novel DAD-SGM to effectively transfer knowledge from self-supervised GNNs into MLPs, an area with limited prior research.
Our work first trains a diffusion-based assistant model that learns a noise prediction function.
It then aligns the noise prediction of teacher outputs with student outputs through the assistant diffusion model.
This approach further improves MLPs in generating high-quality node representations across graphs with diverse characteristics.
We expect our work to offer a scalable and efficient way to analyze large graphs without the need for expensive human annotations.
Extensive experiments demonstrate the effectiveness of DAD-SGM in enhancing the capability and robustness of MLPs in self-supervised graph representation learning. 
However, our work lacks consideration of addressing heterogeneous graphs where nodes and edges have diverse types. 
In future work, we will modify our assistant model to a conditional diffusion model that can model various types of nodes.

\bibliographystyle{ieeetr}
\bibliography{reference2}

\appendix
\subsection{Inductive Node Classification}
\label{appendix}
Real-world deployment often requires classifying nodes in inductive settings, where new nodes appear only at inference time. 
To assess DAD-SGM's effectiveness under these conditions, we conduct a comprehensive experiment following the protocol in NOSMOG~\cite{tian2022nosmog}.
Specifically, we split the node set $\mathcal{V}$ into observed and unobserved subsets ($\mathcal{V} = \mathcal{V}_\text{obs} \cup \mathcal{V}_\text{unobs}$), with $\mathcal{V}_\text{unobs}$ comprising 20\% of the test set.
We have used $\mathcal{V}_\text{obs}$ to train self-supervised teacher models and GNN-to-MLP distillation methods. 
The edges connected to $\mathcal{V}_\text{unobs}$ are removed during training.
For NOSMOG and DAD-SGM, the positional features of $\mathcal{V}_\text{unobs}$ are generated by averaging those of their observed neighbors.
Table~\ref{inductive} presents the inductive performance on five high-homophily datasets, demonstrating that DAD-SGM outperforms all baselines across multiple self-supervised GNN teachers (GAE, GRACE, DGI, CCA-SSG, and GBT). 
These findings confirm that DAD-SGM excels not only in transductive but also in inductive scenarios, indicating its practical potential for real-world graph applications.

\begin{table}[ht]
\centering
\caption{Inductive node‐classification accuracy on graphs with high homophily.
\textbf{Bold} = best, \underline{Underline} = second best.}
\label{inductive}
\resizebox{0.48\textwidth}{!}{%
\begin{tabular}{c|lc|ccccc|c}
\toprule
\multirow{2}{*}{\textbf{Dataset}} &
\multicolumn{2}{c|}{\textbf{Teacher}} &
\multirow{2}{*}{\textbf{GLNN}} & \multirow{2}{*}{\textbf{NOSMOG}} & \textbf{FF} & \multirow{2}{*}{\textbf{HGMD}} & \multirow{2}{*}{\textbf{LLP}} & \textbf{DAD} \\
 & \multicolumn{2}{c|}{\textbf{GNN}} & 
 & \ & \textbf{G2M} & &  & \textbf{SGM} \\
\midrule
\multirow{5}{*}{\textbf{Cora}}
& GAE   & 0.802
  & 0.754 & 0.795 & 0.780 & \underline{0.812} & 0.769 & \textbf{0.821} \\
& GRACE                 & 0.817
  & 0.771 & 0.813 & 0.809 & \underline{0.815} & 0.802 & \textbf{0.822} \\
& DGI                   & 0.796
  & 0.729 & 0.792 & \underline{0.793} & 0.781 & 0.761 & \textbf{0.814} \\
& CCA               & 0.825
  & 0.793 & 0.819 & 0.820 & \underline{0.826} & 0.777 & \textbf{0.837} \\
& GBT                   & 0.814
  & 0.775 & 0.810 & 0.822 & \underline{0.825} & 0.783 & \textbf{0.827} \\
\midrule
\multirow{5}{*}{\textbf{Citeseer}}
& GAE                   & 0.688
  & 0.671 & 0.680 & 0.694 & \underline{0.698} & 0.691 & \textbf{0.703} \\
& GRACE                 & 0.714
  & 0.692 & 0.711 & \underline{0.728} & 0.722 & 0.712 & \textbf{0.730} \\
& DGI                   & 0.697
  & 0.671 & \underline{0.705} & 0.697 & 0.693 & 0.679 & \textbf{0.722} \\
& CCA               & 0.704
  & 0.693 & 0.702 & 0.722 & \underline{0.728} & 0.699 & \textbf{0.729} \\
& GBT                   & 0.719
  & 0.720 & 0.704 & 0.723 & \underline{0.730} & 0.701 & \textbf{0.732} \\
\midrule
\multirow{5}{*}{\textbf{Pubmed}}
& GAE                   & 0.784
  & 0.761 & \underline{0.799} & 0.789 & 0.795 & 0.691 & \textbf{0.805} \\
& GRACE                 & 0.784
  & 0.795 & 0.797 & 0.802 & \underline{0.806} & 0.801 & \textbf{0.807} \\
& DGI                   & 0.788
  & 0.788 & 0.788 & 0.795 & \underline{0.799} & 0.693 & \textbf{0.820} \\
& CCA               & 0.792
  & \underline{0.799} & 0.798 & 0.788 & 0.781 & 0.687 & \textbf{0.802} \\
& GBT                   & 0.788
  & 0.776 & 0.781 & \underline{0.786} & 0.765 & 0.684 & \textbf{0.803} \\
\midrule
\multirow{5}{*}{\textbf{CS}}
& GAE                   & 0.914
  & 0.917 & \textbf{0.933} & 0.918 & 0.924 & 0.931 & \textbf{0.933} \\
& GRACE                 & 0.924
  & 0.916 & 0.916 & 0.920 & 0.917 & \textbf{0.926} & \underline{0.923} \\
& DGI                   & 0.920
  & 0.916 & 0.916 & 0.898 & \underline{0.919} & 0.904 & \textbf{0.921} \\
& CCA               & 0.918
  & 0.912 & 0.898 & 0.885 & \underline{0.915} & 0.907 & \textbf{0.935} \\
& GBT                   & 0.919
  & \underline{0.913} & 0.905 & 0.878 & 0.886 & \underline{0.913} & \textbf{0.935} \\
\midrule
\multirow{5}{*}{\textbf{Physics}}
& GAE                   & 0.948
  & \underline{0.941} & 0.940 & 0.939 & \underline{0.941} & \underline{0.941} & \textbf{0.943} \\
& GRACE                 & 0.951
  & 0.947 & 0.952 & 0.948 & 0.950 & \textbf{0.953} & \textbf{0.953} \\
& DGI                   & 0.948
  & 0.941 & \underline{0.943} & 0.940 & 0.940 & 0.938 & \underline{0.947} \\
& CCA               & 0.953
  & \underline{0.943} & \underline{0.943} & 0.935 & \underline{0.943} & 0.938 & \textbf{0.945} \\
& GBT                   & 0.951
  & 0.942 & \underline{0.943} & 0.937 & 0.942 & \underline{0.943} & \textbf{0.944} \\
\bottomrule
\end{tabular}
}
\end{table}

\subsection{Impact of Noise Prediction Quality on Knowledge Transfer}
\begin{figure}[!t]
    \centering
        \centerline{\includegraphics[width=0.53\textwidth]{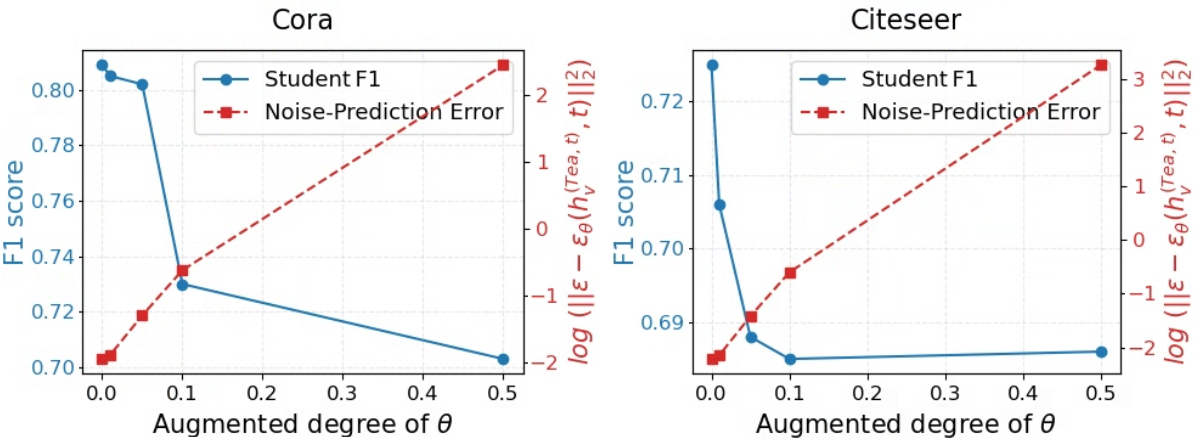}}
        \caption{Impact of the noise-prediction quality on knowledge transfer.}
        \label{reviewer1-figure2}
    \vspace{-1em}
\end{figure}

Our diffusion-based teacher assistant (TA) improves knowledge transfer by learning and then utilizing a noise-prediction function $\epsilon_\theta$.
In this section, we investigate how the quality of noise prediction influences knowledge transfer. 
To this end, we disturb our TA's parameters $\theta$ by injecting zero-mean Gaussian noise with standard deviations $\sigma\in\{0,0.01,0.05,0.1,0.5\}$.
For each $\sigma$, we distill a student MLP and measure the noise prediction error $||\epsilon-\epsilon_\theta(x_t, t)||_2^2$.
Figure \ref{reviewer1-figure2} plots the student F1 scores (left axis) along with $log(||\epsilon-\epsilon_\theta(x_t, t)||_2^2)$ (right axis) on Cora and Citeseer.
For both datasets, student performance increases as the noise prediction error decreases, exhibiting a strong negative correlation. 
These results confirm that precise noise prediction is essential for effective knowledge transfer from the diffusion‑based TA to the student.

\subsection{Other Semi-Supervised Teachers}
{\color{black}{
We also evaluate our approach in distilling two other supervised GNN teachers: GAT and APPNP.
Here, we use a 3-layer GAT with 4 heads.
For APPNP, we set the propagation step $K$ and teleport rate $\alpha$ to 10 and 0.1, respectively.
Table \ref{tab:gat_teacher} and \ref{tab:appnp_teacher} show that DAD-SGM matches or outperforms existing techniques without using labels or teacher logits.
This confirms that our method remains highly effective when the teacher GNN is trained in a supervised setting.}}

\begin{table}[!ht]
\centering
\caption{Node Classification performance with a GAT teacher.}
\label{tab:gat_teacher}
\resizebox{\linewidth}{!}{%
\begin{tabular}{lc|ccccc}
\toprule
 & \textbf{GAT} & \multirow{2}{*}{\textbf{GLNN}} & \multirow{2}{*}{\textbf{NOSMOG}} & \textbf{FF} & \multirow{2}{*}{\textbf{HGMD}} & \textbf{DAD‑SGM} \\
 & \textbf{(Teacher)} &  &  & \textbf{G2M} &  & \textbf{(Ours)} \\
\midrule
\textit{Use Label?} & \ding{51} & \ding{51} & \ding{51} & \ding{51} & \ding{51} & \ding{55} \\
\textit{Used Info.} &  & logits & logits + repr & logits & logits & repr \\
\midrule
Cora        & 0.790 & 0.804 & 0.804 & \underline{0.809} & 0.801 & \textbf{0.819} \\
Citeseer    & 0.695 & 0.717 & 0.718 & \textbf{0.722} & 0.704 & \underline{0.720} \\
Pubmed      & 0.791 & 0.804 & 0.794 & \underline{0.810} & 0.809 & \textbf{0.813} \\
CS          & 0.932 & 0.913 & 0.914 & 0.908 & \underline{0.916} & \textbf{0.917} \\
Physics     & 0.935 & 0.944 & 0.946 & \underline{0.951} & 0.949 & \textbf{0.954} \\
\bottomrule
\end{tabular}}
\end{table}

\begin{table}[!ht]
\centering
\caption{Node Classification performance with an APPNP teacher.
}
\label{tab:appnp_teacher}
\resizebox{\linewidth}{!}{%
\begin{tabular}{lc|ccccc}
\toprule
 & \textbf{APPNP} & \multirow{2}{*}{\textbf{GLNN}} & \multirow{2}{*}{\textbf{NOSMOG}} & \textbf{FF} & \multirow{2}{*}{\textbf{HGMD}} & \textbf{DAD‑SGM} \\
 & \textbf{(Teacher)} &  &  & \textbf{G2M} &  & \textbf{(Ours)} \\
\midrule
\textit{Use Label?} & \ding{51} & \ding{51} & \ding{51} & \ding{51} & \ding{51} & \ding{55} \\
\textit{Used Info.} &  & logits & logits + repr & logits & logits & repr \\
\midrule
Cora        & 0.819 & 0.826 & 0.821 & \underline{0.833} & 0.832 & \textbf{0.836} \\
Citeseer    & 0.704 & 0.714 & 0.711 & \underline{0.727} & 0.712 & \textbf{0.734} \\
Pubmed      & 0.777 & 0.790 & 0.790 & 0.790 & \underline{0.798} & \textbf{0.811} \\
CS          & 0.930 & 0.904 & 0.914 & 0.909 & \textbf{0.915} & \textbf{0.915} \\
Physics     & 0.937 & \underline{0.947} & 0.943 & 0.945 & 0.945 & \textbf{0.951} \\
\bottomrule
\end{tabular}}
\end{table}

\begin{IEEEbiography}[{\includegraphics[width=1in,height=1.25in,clip,keepaspectratio]{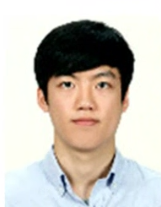}}]{Seong Jin Ahn}
Seong Jin Ahn received the B.S. degree in Mathematics from Sogang University, Seoul, Republic of Korea, in 2019, and the M.S. degree in Computer Science from the School of Computing, KAIST, Daejeon, Republic of Korea, in 2021. 
He is currently a Ph.D. candidate with the School of Computing, KAIST. His research interests include diffusion models, graph learning, and data imputation.
\end{IEEEbiography}

\begin{IEEEbiography}[{\includegraphics[width=1in,height=1.25in,clip,keepaspectratio]{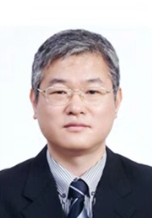}}]{Myoung-Ho Kim}
Myoung-Ho Kim received the B.S. and M.S. degrees from Seoul National University, Seoul, Republic of Korea, in 1982 and 1984, respectively. 
He received the Ph.D. degree from Michigan State University, East Lansing, MI, USA. 
He is currently a Professor with the School of Computing, KAIST, Daejeon, Republic of Korea. 
His research interests include database systems, data mining, and multimedia data.
\end{IEEEbiography}

\end{document}